\documentclass{article}

\usepackage{microtype}
\usepackage{graphicx}
\usepackage{booktabs} %

\usepackage{color, colortbl}
\usepackage{enumitem}
\usepackage{subcaption}
\usepackage{caption}

\PassOptionsToPackage{hyphens}{url}
\usepackage{hyperref}

\usepackage{multirow}
\usepackage{soul}
\usepackage{makecell}

\usepackage[accepted]{icml2024}

\usepackage{amsmath}
\usepackage{amssymb}
\usepackage{mathtools}
\usepackage{amsthm}

\usepackage[capitalize,noabbrev]{cleveref}

\theoremstyle{plain}

\theoremstyle{definition}

\theoremstyle{remark}

\usepackage[textsize=tiny]{todonotes}

\begin{document}

\twocolumn[
\icmltitle{CARTE: Pretraining and Transfer for Tabular Learning}

\begin{icmlauthorlist}
\icmlauthor{Myung Jun Kim}{soda}
\icmlauthor{Léo Grinsztajn}{soda}
\icmlauthor{Gaël Varoquaux}{soda,probabl}
\end{icmlauthorlist}

\icmlaffiliation{soda}{SODA Team, Inria Saclay, France}
\icmlaffiliation{probabl}{Probabl.ai, France}

\icmlcorrespondingauthor{Myung Jun Kim}{myung.kim@inria.fr}

\icmlkeywords{Machine Learning, ICML}

\vskip 0.3in
]

\printAffiliationsAndNotice{}  %

\begin{abstract}
Pretrained deep-learning models are the go-to solution for images or text. However, for tabular data the standard is still to train tree-based models. Indeed, transfer learning on tables hits the challenge of \emph{data integration}: finding correspondences, correspondences in the entries (\emph{entity matching}) where different words may denote the same entity, correspondences across columns (\emph{schema matching}), which may come in different orders, names... We propose a neural architecture that does not need such correspondences. As a result, we can pretrain it on background data that has not been matched. The architecture --CARTE for Context Aware Representation of Table Entries-- uses a graph representation of tabular (or relational) data to process tables with different columns, string embedding of entries and columns names to model an open vocabulary, and a graph-attentional network to contextualize entries with column names and neighboring entries. An extensive benchmark shows that CARTE facilitates learning, outperforming a solid set of baselines including the best tree-based models. CARTE also enables joint learning across tables with unmatched columns, enhancing a small table with bigger ones. CARTE opens the door to large pretrained models for tabular data.
\end{abstract}

\section{Introduction}

The wide availability of pre-trained models has greatly facilitated machine learning for various data modalities, for example with images \citep{simonyan2015very} or texts \citep{devlinBERTPretrainingDeep2019}. These models can be downloaded from model hubs, embarking a lot of implicit information and transformations that unleashes the power of deep learning even on small datasets. This paradigm has led to the revolution of foundation models \citep{bommasani2021opportunities} such as large language models \citep{touvron2023llama}. But this revolution has not happened for tabular data despite its huge importance for enterprise and institutional data. One roadblock is that integrating data from tables in the wild is often difficult, sometimes impossible. Different tables might not have any related data and when they do, data integration is a whole field of database research \citep{doan2012principles}. One might need to solve correspondence across columns --\emph{schema matching}-- or across data sources with different naming conventions for entries --\emph{entity matching}.
For lack of matching schemas and entities, pretraining across tables in the wild has not been possible. Without pretraining, deep learning is less practical, and tree-based methods are often preferable \citep{grinsztajnWhyTreebasedModels2022}.

Here we introduce a learning architecture that learns across tables without schema and string matching. The key is to represent tables with a graph and all symbols with embeddings (for column names and table entries). The architecture, dubbed CARTE (Context-Aware Representation of Table Entries), is pretrained on a large knowledge base, to capture information on a vast amount of entities and relations. It can then be fine-tuned on a given downstream task, helping learning even in few shot settings. It can also be used for joint learning across multiple tables, enriching a target table with weakly related sources. 
CARTE brings a sizable performance gain, outperforming markedly a set of 42 solid baselines (including the best tree-based methods and various feature engineering).
It benefits particularly tables with string entries, frequent in applications but seldom present in machine learning benchmarks.

Section \ref{sec:related} presents related work; \autoref{sec:carte} describes the CARTE architecture and training procedures; and \autoref{sec:experiments} provides an extensive empirical study across many tabular datasets, benchmarking the settings of a single downstream table as well as multiple related ones.

\section{Related Works}\label{sec:related}

\paragraph{Tabular deep learning}
Tables are central to many applications. As a result, numerous deep learning methods tailored for this modality have been proposed \citep{abutbulDNFNetNeuralArchitecture2020a, arikTabNetAttentiveInterpretable2020,popovNeuralObliviousDecision2019, gorishniyRevisitingDeepLearning2023, somepalliSAINTImprovedNeural2021a}. However, they typically under-perform tree-based methods \citep{grinsztajnWhyTreebasedModels2022, shwartz-zivTabularDataDeep2021b,gardnerSubgroupRobustnessGrows2022}. While \citet{mcelfreshWhenNeuralNets2023} argue that neural networks perform well on certain types of tables, and promising architectures are continuously published \citep{gorishniyTabRTabularDeep2023, chenExcelFormerNeuralNetwork2023}, the difficulty of improving over tree-based methods suggests that deep learning must bring something more to the fight, such as background knowledge.

\paragraph{Transfer learning for tabular data} 
Transfer learning mostly focuses on the ``conventional'' settings of transferring across datasets with the same features, \emph{i.e.} columns. \citet{somepalliSAINTImprovedNeural2021a} demonstrates pre-training on a larger unlabeled version of the table, while \citet{levinTransferLearningDeep2023} argues that transfer learning bridges the gap between deep learning and tree-based models when there are few data points but large related datasets, as in their medical setting. They consider new or missing features in the downstream table, but require exact matching between most features.
 
XTab \citep{zhu2023xtab} can work on tables with different columns using data-specific featurizers that map instances to the same dimension followed by federated learning on the common block. However, they did not outperform tree-based models \citep[CatBoost,][]{dorogushCatboost2018}.
Transtab \citep{wang2022transtab} also vectorizes each row of tables into an embedding space to learn across tables, demonstrating data accumulation across multiple clinical trials to outperform baselines including XGBoost \citep{chen2016xgboost}. These approach benefit from a pool of tables in the subtopic, but it is not clear if they can be adapted to build pretrained models for a wide set of applications.

\paragraph{Pretrained models for tabular data} 
TabPFN \citep{hollmannTabPFNTransformerThat2023} made headway in pre-training models for tabular learning: it uses a transformer model pre-trained on large amounts of synthetic data to capture the inductive biases of tabular data, leading to strong performance on small datasets, though it has no dedicated handling for categorical columns, a challenge of tables where trees historically shine. Large language models (LLMs) can also work as pretrained models for tabular data. In TabLLM \citep{hegselmannTabLLMFewshotClassification2023}, tabular data are represented as a set of tokens which are leveraged to fine-tune an LLM. However, the difficulty of handling numerical values in LLMs makes them a suboptimal choice compared to trees or TabPFN.

\paragraph{Discrete entries}
One challenge of tables --much more tackled by the database literature than the machine learning one-- is that many of the entries are discrete, represented as strings. \citet{cerdaEncodingHighcardinalityString2022} created string-based representations that facilitate learning. The {\tt TableVectorizer} in \citet{skrub2024} uses these heuristically to turn columns of different types into numerical matrices well suited for learning.
 KEN \citep{cvetkov2023relational} is another approach to embed table entities closer to our goals of pretraining. It provides embeddings of all entities in a knowledge graph, capturing the information in a source such as Wikipedia. These embeddings facilitate learning, but the challenge is that each entry of a column must be linked to a Wikipedia entry, an \emph{entity matching} task.

\paragraph{Data integration} Traditional statistical models need data assembled in a single consistent table, a task tackled by the data integration literature \citep{doan2012principles}. Finding correspondences between columns across data sources is known as \emph{schema matching}. \emph{Entity matching} is a challenge common to data integration and natural language processing (NLP), where a string must be linked to an \emph{entity}: a unique concept. For instance ``Davinci'' may denote the historical figure ``Leonardo da Vinci'', but also OpenAI ``Text-Davinci-003'' GPT3 API. Entity matching must be robust to string variations, but most importantly it must account for the context in which an entity appears to disambiguate potential matches. In NLP, pretrained attention-based models, such as BERT \citep{devlinBERTPretrainingDeep2019}, have been crucial to capture the corresponding context.

These pretrained language models also are useful on tables to automate data normalization and integration tasks with few manually-supplied examples \citep{narayan2022can}. Deep learning, and recently attention-based models, is progressing on tasks to structure databases, \emph{e.g.,} column typing, entity linking \citep{hulsebos2019sherlock,dengTURLTableUnderstanding2020}.

\bigskip%
\vfill

We are interested in a different problem: rather than explicit matching, at the entity or schema level, we aim to capture only implicit data structure and integration to enhance downstream machine learning task without any manual operation such as finding related sources. 
The problem is timely: researchers sharing this vision are assembling large corpora of tables \citep{hulsebosGitTablesLargeScaleCorpus2023,eggertTabLibDataset627M2023}. Yet, the small scale of most tables available and the variability between tabular datasets has so far made this vision elusive.

\section{The CARTE Model to Learn Across Tables}\label{sec:carte}
The ability of CARTE to learn across tables stems from the combination of two elements: a novel representation of table entities with graphs and a deep neural network architecture that captures the context that reside within a table. In particular, the former endows synchronization of multiple tables to the same graph domain, which makes pretraining on formerly unmatched background data possible. Moreover, the context-aware deep neural network trained on broad spectrum of knowledge can readily spread the background information to downstream tasks at our hands. In this section, we introduce CARTE with detailed implementations. 

\subsection{Graph Representation of Table Entities}\label{subsec:graphlet_construction}
\begin{figure}
    \includegraphics[width=\linewidth]{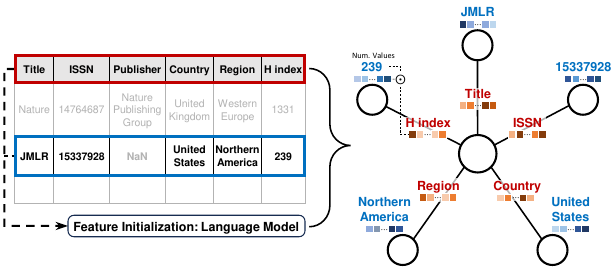}%
    \caption{\textbf{Graphlet representation of tabular entities.} From a table, CARTE represents each row as a star-like graph. Excluding for missing values, the leaf-nodes and the edges are annotated by the cell values and their corresponding column names. Then, CARTE initializes the features of each with a language model. The nodes of numerical values are initialized by the elementwise product with its corresponding column feature. For the center node, it is initially set as the average of the leaflets. It will later act as a readout that captures the overall information of the graphlet.%
    \label{fig:graphlet_construction}
    }
\end{figure}

The graph representation is crucial for facilitating the generalization of table entities. In general, a graph, $G$, consists of nodes and edges, where the former denote the entities and the latter denote the relations between the nodes. Graphs are useful for capturing relational information between entities, and graph deep learning is promising for relational databases \citep{fey2023relational}. CARTE considers each of the row as a small graph, as shown in \autoref{fig:graphlet_construction}. From a table with $k$ columns, CARTE represents each of the $i$-th instance as a graph, $G_{i}(X, E)$, where the components $X$ and $E$ denote the node and edge features, respectively, embeddings in $\mathbb{R}^d$. The structure of $G_{i}(X, E)$ is a star-like graph with $k-p_{i}$ leaf-nodes, with $p_{i}$ as the number of columns with missing values for row $i$. On the resulting graphlet, each of the leaf-nodes are annotated by the cell values and their corresponding column names. To make these graphlets viable inputs for neural networks, we initialize $X$ and $E$ by using a language model. For categorical values and column names, CARTE simply places a $d$-dimensional embedding that is generated from a language model. For numerical values, the features are initialized by the product of its value with the embedding of the corresponding column name. For instance, the node feature $X_{(239)}$ in \autoref{fig:graphlet_construction} is equal to $239 \times E_{\text{(H index)}}$. Lastly, the center node is initialized with the mean of the leaflets, and will later serve as the readout node that captures the overall information of the graph.

This design serves several purposes. First it represents context: for tabular data, an entry is best interpreted accounting for its column's name. In \autoref{fig:graphlet_construction}, for example, it would be difficult to grasp the row (blue-box) only with the entries of `JMLR', `15337928', `$239$'. The column names `Title', `ISSN', and `H index' clarifies that it is an instance of a journal. CARTE represents context in tables through the nodes and edge, exposed to its neural network architecture.
This representation also bridges tables with different column order, or more generally different columns.

Second, CARTE uses language models on non-numerical entries, such as strings, categories, and names. Thus, the graph transformation in CARTE does not require any intervention on discrete entries, as opposed to typical data preprocessing or cleaning (deduplication, categorical encodings) used on strings. Moreover, CARTE works with an open set of vocabularies. Problems of typos or wordings of the same meaning, such as `North America' to `Northern America', are readily resolved for CARTE.

Together, these features of the proposed graph representation enable generalization across heterogeneous tables. CARTE's graph transformation puts in the same  graph domain instances from different tables, without requiring any schema matching for columns or entity matching for entries. Thus, learning process can operate across many tables, which opens the door for pretraining or transfer.

\subsection{Pretrained Model from a Large Knowledge Base}
\begin{figure}
    \includegraphics[width=\linewidth]{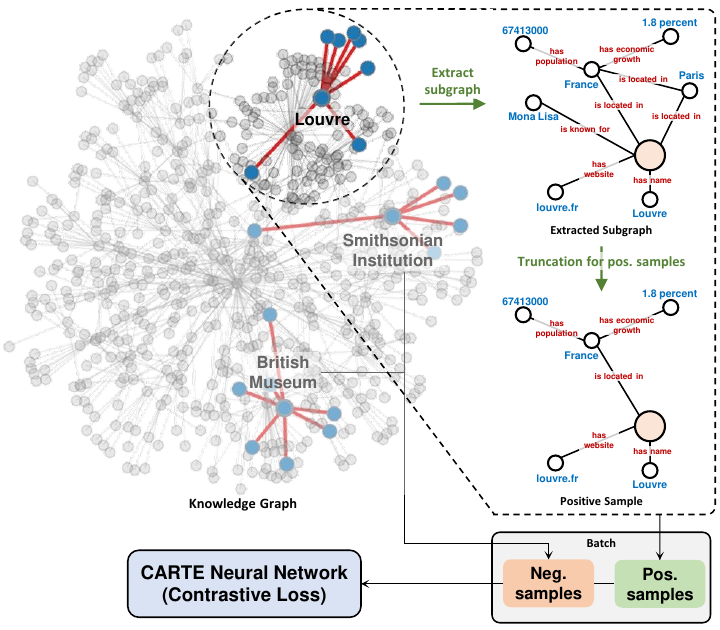}%
    \caption{\textbf{CARTE pretraining process.} From a large knowledge graph, CARTE begins by constructing graphlets and their positives variants. The extracted samples are then fed into the CARTE neural network and trained with a self-supervised scheme. The neural network learns to aggregate information within the graphlets, which reflect the combination of table entries across columns (edges).%
    \label{fig:pretrain_process}
    }
\end{figure}

CARTE is pretrained on YAGO3 \citep{mahdisoltani2013yago3}, a large knowledge base built from Wikidata and other sources that contain facts about real-world. YAGO stores information as a knowledge graph, which is a collection of triplets \emph{(head, relation, tail)}. For instance, the triplet \emph{(Louvre, is located in, Paris)} from \autoref{fig:pretrain_process} would be a sample that we can find in YAGO. Our current version of YAGO contains over $18.1$ million triplets of $6.3$ million entities.

In this subsection, we describe the pretraining process of CARTE, summarized in \autoref{fig:pretrain_process}. From the knowledge graph, we first extract small graphlets of entities suitable as inputs for CARTE. For self-supervised learning with a contrastive loss, we add to the batch truncated versions of selected graphlets. Through the training process, CARTE learns to aggregate information based on the given context. 

\paragraph{Graphlets for pretraining} From the large knowledge graph of YAGO, we construct small graphlets of the entities that can be used as inputs for CARTE. To construct a suitable graphlet for an entity, we first extract its subgraph within a user-specified $k$-hop relations. To resemble the structure outlined in \autoref{fig:graphlet_construction} while benefiting from additional information through multiple hops, we set $k=2$, but restricting the maximum number of $1$-hop and $2$-hop relations to $100$ and $10$ respectively. Graphlets from tables (\autoref{fig:graphlet_construction}) have as center node a token for the row, while the knowledge-graph procedure could use the entity name (for instance `Louvre'). To avoid a difference, we use a token as a center node with an additional neighbor which is comprised of the name as its node and `has name' as its relation. Finally, as in \autoref{subsec:graphlet_construction}, we initialize node and edge features using FastText embeddings \citep{mikolov2017advances} as the language model. 

\paragraph{Batch samples} To construct a batch sample of size $N_{b}$, we first select which of the YAGO entities to include, generating the corresponding graphlets. For this, we sample $0.9$ of $N_{b}$ from entities with $6$ or more $1$-hop relations and the remaining $0.1$ from the other subset. The main reason for such sampling scheme is that a large portion of entities in YAGO only have one or two $1$-hop relations, while tabular data typical has more (more columns). Moreover, the value $6$ was selected so that the rough median of $1$-hop relations in the batch samples is $15$. To enable the self-supervised contrastive loss, we include positive samples, which are simply truncations of original graphlets: deleting a random fraction (varying from $0.3$ to $0.7$) of the edges. \autoref{fig:pretrain_process} gives an exemple graphlet of `Louvre' and its positive. 

\begin{figure}
    \includegraphics[width=\linewidth]{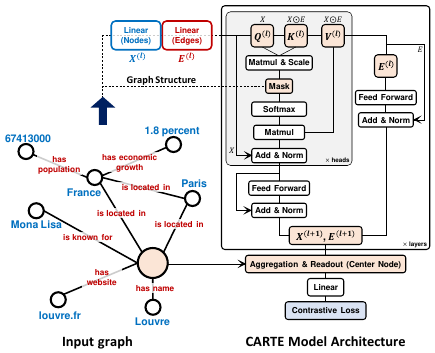}%
    \caption{\textbf{CARTE architecture} The inputs of CARTE are graphs that contain node ($X$) and edge ($E$) features, both used in self-attention layers (shown in grey). The attention layers update node features using the context embodied with the edge information; the graph structure of the input is reflected by attention masks. The Aggregate \& Readout layer consists of the attention layer (without the edge update) followed by feature extraction on the center node. The outputs are then processed for the contrastive loss.%
    \label{fig:carte_nn_architecture}
    }
\end{figure}

\paragraph{Model architecture} \autoref{fig:carte_nn_architecture} depicts the model structure of CARTE. At its basis, CARTE takes the classical Transformer encoder model of \citet{vaswani2017attention}, and adapts to a graph attentional network. A key component in CARTE's architecture is a self-attention layer computing attention from both node and edge features. In graph models, attention modulates the importance of neighbors for a given node of interest \citep{velickovic2017graph}. For table entries, it translates to the importance of the entries for a given instance with the context supplemented by the column information.

We now detail on CARTE's attention mechanism used to capture context and relations. For consistent notations, we write vectors with an arrow on top $\vec{A}$, matrices in bold $\mathbf{A}$, and scalars $A$. To ease reading, we present a single-head attention layer, but it can easily be extended to multi-head schemes of concatenating or averaging the attention outputs.

For a graph with $N$ nodes, let $\vec{X}^{(l)}_{i} \in \mathbb{R}^{d}$ denote the feature of node $i$ and $\vec{E}^{(l)}_{ij}\in \mathbb{R}^{d}$ the feature of the edge directed from node $j$ to $i$. By design, graphlets for CARTE always hold the center node, which we denote with an index $i=1$. The representation from the attention layer is a function of query, key, and value, crucial elements to account for context. The query is the vectors corresponding to our values of interest: the nodes. Thus, we take the conventional approach and parameterize it with solely the node information. On the other hand, the key-value pairs should convey the elements that the neighboring nodes can offer. Therefore, we add edge information in the corresponding parameterization. With this in mind we set the three components as\footnote{Here, we omit the superscripts indicating layers for $A_{ij}$, $e_{ij}$, and projection weights for $Q$, $K$, $V$ for clarity of presentation.}:
\begin{flalign}
\text{Query:}&&
\vec{Q}_{i} & = \vec{X}^{(l)}_{i}\cdot \mathbf{W_{Q}} &&\\
\text{Key:}&&
\vec{K}_{ij} & = (\vec{X}^{(l)}_{i}\odot \vec{E}^{(l)}_{ij})\cdot \mathbf{W_{K}} &&
\label{eq:key}%
\\
\text{Value:}&&
\vec{V}_{ij} & = (\vec{X}^{(l)}_{i}\odot \vec{E}^{(l)}_{ij})\cdot \mathbf{W_{V}}&&
\label{eq:value}%
\end{flalign}
where $\odot$ denotes the element-wise multiplication and $\mathbf{W_{Q}}$, $\mathbf{W_{K}}$, and $\mathbf{W_{V}}$ are trainable weights that reside on $\mathbb{R}^{d\times d}$. Here, the choice of element-wise product is motivated from a knowledge graph embedding technique in \citet{balazevic2019multi, cvetkov2023relational}. These works showed that modeling relations (\emph{i.e.} column name) as element-wise multiplication on the node vectors works best, compared to \emph{e.g.,} vector additions. Following the scaled dot-product attention with the above three equations, the attention score of node $j$ from node $i$, $A_{ij}$, is derived as: 
\begin{equation}
A_{ij} = \frac{\exp{(e_{ij})}}{\sum_{k\in\mathcal{N}_i}\exp{(e_{ik})}}, \,\,\text{where}\,\, e_{ij} = \frac{\vec{Q}_{i}\cdot \vec{K}_{ij}^{T}}{\sqrt{d}}
\end{equation}
where the calculation of $A_{ij}$ only takes the sum with respect to the connected neighbors of node $i$. This corresponds to a masking step, which takes the graph structure of the input. Accounting for the relation type (\emph{i.e.} column name) in the attention scores ($\vec{E}$ in eq\,\ref{eq:key} and \ref{eq:value}) is important for proper re-contextualization of the entries, that is to capture their meaning. For instance, an entry ``\emph{George Bush}'' may denote the 41\textsuperscript{st} or 43\textsuperscript{rd} US presidents, an aircraft carrier... The ambiguity is raised by the relation (``\emph{George Bush}'', ``\emph{son of}'', ``\emph{George Bush}''), however capturing fully the information does require knowing the nature of the relation, as ``\emph{father of}'' would lead to a different entity resolution.
Ablations reveal the importance of attention (Appendix \ref{app:ablation_attention}).

Outputs of the attention layers are, for nodes and edges:
\begin{flalign*}
\text{transformed entry} &&
\vec{X}_{i}^{(l+1)} & = \sigma_{X}\bigl(\sum_j A_{ij}\cdot \vec{V}_{ij}\bigr) 
&&\\
\text{transformed relation} &&
\vec{E}_{ij}^{(l+1)} & = \sigma_{E}(E^{(l)}_{ij}\cdot \mathbf{W_{E}})
&&
\end{flalign*}
where $\sigma$ denote the appropriate consecutive operations (see \autoref{fig:carte_nn_architecture}). The final layers consist of the attention layer without the edge update, followed by the readout layer that extracts the representation of the center node. For pretraining, the outputs are then processed for the contrastive loss. Appendix \ref{app:pretrain} details model specification and training. 

\paragraph{Contrastive loss} For the self-supervised contrastive loss, we adapt the framework of \citet{chen2020simple}. The original graphlet and one truncation are set as positives while the other graphlets in the batch are considered as negatives. The learning loss is then based on the cosine similarity of the network outputs, fed in the InfoNCE loss \citep{oord2018representation}.

\subsection{Fine-tuning for Downstream Tasks}

For a given downstream task, fine-tuning CARTE proceeds by reusing only part of the pretrained architecture (as shown in \autoref{fig:carte_nn_architecture}): the initial layers for nodes and edges (blue and red blocks) and the `Aggregation \& Readout' layer. Though such simplification differs from many fine-tuning approaches, it stems from the behavior of graph-neural networks.
Indeed, downstream table entities form simpler graphs than during pre-training. First they are star-like (\autoref{fig:graphlet_construction}). 
Second, the downstream tables contain less variability in graph structures and less cardinality of discrete variables compared to YAGO.
Too deep an architecture risks washing out discriminant characteristics in the output representations \citep[the over-smoothing problem,][studied in Appendix \ref{app:oversmoothing}]{chen2020measuring, rusch2023survey}. Therefore, we use a convention common in graph models: setting the number of attention layers as the maximum $k$-hop relation, here $k=1$. For the final classifier, we simply attach the linear layers. With the base model for fine-tuning, we consider two different settings of downstream inference. 

\paragraph{Inference on single tables} This is the well-known setting in which we are given a single table with a target variable to predict. Before transforming table entities into graphs, we preprocess numerical variables with a power transform \citep{yeo2000new}. The power transform has shown to be effective in several works \citep[\emph{e.g.,}][]{hollmannTabPFNTransformerThat2023, cvetkov2023relational}, and likewise, gives stability to the fine-tuning process of CARTE. Moreover we employ a bagging strategy \citep{breiman1996bagging}, in which different models, based on different train-validation splits used for early-stopping, are trained. The prediction outputs are calculated as the average of the outputs from each model.

\paragraph{Transfer from one source table to a target} We also use CARTE in transfer learning settings where we are given a source table $X_S$ that can aid predictions on our target table $X_T$. Importantly, the source table may have larger train samples than the target. We fine-tune CARTE on both tables jointly. The graph representation enables such joint fine tuning without correspondences in the columns; however, we do need to have similar outcomes $y_S$ and $y_T$ on both tables. The source outcomes $y_S$ are transformed to match the first moment of the target outcome $y_T$ using as above a power transform \citep[note that here we use the inverse transform]{yeo2000new}. If the target and source table differ on the classification / regression nature of the outcome, we adapt $y_S$ as the following: for a classification target $y_T$, we binarize regression outcomes in the source table, and for a regression target $y_T$, we use binary classification outcomes of the source table, encoded as $\{0, 1\}$ and standard scaled. We then proceed to fine tune CARTE by drawing batches with a fixed proportion of rows from the target and source tables (we use a batch size of 64, 8 of which come from the target). We use early stopping on a validation set of the target table, and still rely on the bagging strategy of building multiple learners on different random validation sets and averaging the predictions. Often, early stopping kicks in before all the data points of the source have been covered. This prevents overfitting the source data, which may be less important than the target data for predicting $y_T$. We use the hyperparameters selected in the single-table setting.

As we choose source tables quite loosely from weakly-related data, the resulting pairwise learner may not actually improve upon the single-table learner if the source table does not bring in enough related information. We thus combine the pairwise learner with the single-table learner  ensembling the predictors by combining their output with a softmax. The weights of the softmax are computed using the prediction score computed in the internal validation set of these learners, but divided by the standard deviation across the learners to set the temperature of the softmax.

\paragraph{Joint learning across multiple tables} The key to transfer learning, as above, is finding the right source table. If we have multiple tables from a given domain or institution, CARTE can be adapted to use them all, finding the most useful information for transfer. In these settings we are given a target table $X_{T}$ and a set of source tables ${X_{S,1}...X_{S,m}}$. We proceed to build individual learners: first the single-table learner on $X_T$, then each pair\footnote{To limit computation cost, we do not explore the full combinatorials of source tables.} of one source table $X_{S,i}$ and the target table $X_T$ using the pairwise joint learner described above. Here again, not every pairwise learning brings the same amount of useful information. Thus, to find the optimal combination of datasets, we use the same strategy as above of ensembling all the pairwise predictors as well as the single-table predictor. As a consequence, if all source tables lead to predictors that work as well, they are combined with equal weights, but if one source dominates, the prediction is anchored on this one.

\section{Experimental Study}\label{sec:experiments}

\subsection{Experimental Setup}

\paragraph{Datasets} We use 51 tabular learning datasets, all with an associated learning task --40 regressions and 11 classification--, gathered across multiple sources, mainly from previous machine learning studies and kaggle competitions. They cover a variety of topics of society and businesses: accidents, elections, remunerations, food, restaurants, etc. We select datasets representative of modern data science applications: tables with meaningful columns and discrete entries (\autoref{tab:data_specification}), unlike many datasets from \citet{uci}. Appendix \ref{app:datasets} gives the specific list of datasets.

\paragraph{Baselines} We evaluate different baselines with the following abbreviations (specific details on experiment settings and hyper-parameter tuning are presented in Appendix \ref{app:exp_settings}):

\begin{description}[itemsep=1pt, parsep=1pt, topsep=1pt]
    \item[CatBoost] \citep{dorogushCatboost2018} A gradient-boosted trees package commonly used to learn on tables. We treat text features as categorical, encoded by CatBoost's categorical encoding, an improved version of target encoding \citep{micci2001preprocessing}.
    \item[TabVec] The TableVectorizer from the Skrub package \citep{skrub2024} to encode tables that contain string entries into numerical arrays. Columns with low cardinality (number of categories) are one-hot encoded while those with high cardinality are encoded using the Gamma-Poisson encoder from skrub, introduced in \citet{cerdaEncodingHighcardinalityString2022}, which extracts latent categories from substrings. For non tree-based models, missing values are imputed with the mean for numerical features, and treated as another category for categorical features. For neural network models, minmax scale is applied to set values between zero and one.
    \item[XGB, HGB, and RF] Tree-based models: XGBoost \citep{chen2016xgboost}, HistGradientBoosting and RandomForest \citep[from scikit-learn, ][]{pedregosa2011scikit}.
    \item[MLP and ResNet] The classical Multilayer Perceptron (MLP) and its extension with additional layernorm/batchnorm and skip-connections (ResNet).
    \item[Ridge and Logistic] Linear models, Ridge and Logistic regression for regression and classification tasks.
    \item[\textbf{S-LLM}] Inspired by TabLLM  \citep{hegselmannTabLLMFewshotClassification2023}, we investigate encoding each row of a table with a large language model (LLM). We represent each row as a sentence, and encode them with {\tt intfloat/e5-small-v2} \citep{wang2022text} from HuggingFace. Unlike TabLLM, however, the encoded table is passed to the XGB estimator to enable learning for both regression and classification. For numerical entries, we investigate either concatenating them as additional features outside of the LLM (\textbf{CN}), or passing them as strings to the LLM (\textbf{EN}).
    \item[\textbf{TabPFN}] \citep{hollmannTabPFNTransformerThat2023} is a transformer model pretrained on synthetic data to generate predictions for new (small) datasets in one forward pass. We treat the text features as categorical and encode them with a target encoder \citep{micci2001preprocessing}.
\end{description}

\begin{figure*}[!t]
    \includegraphics[width=0.5\linewidth]{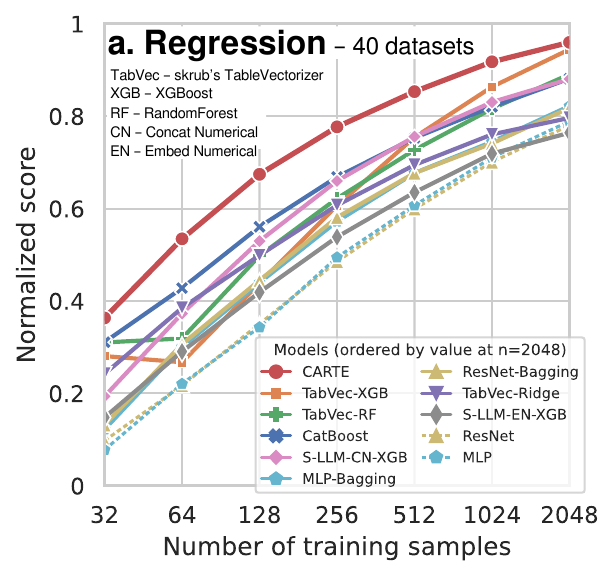}%
    \includegraphics[width=0.5\linewidth]{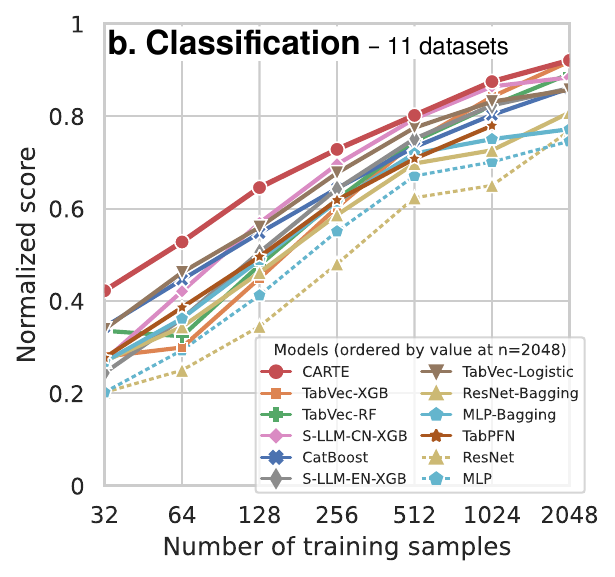}%
        \\%
    \includegraphics[width=0.5\linewidth]{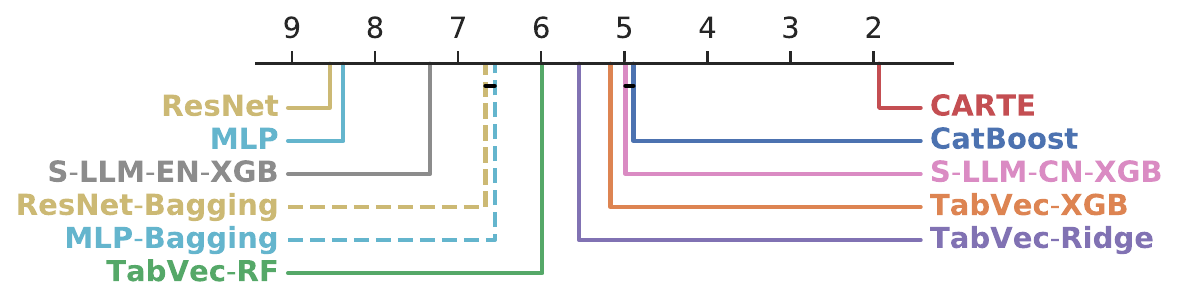}%
    \includegraphics[width=0.5\linewidth]{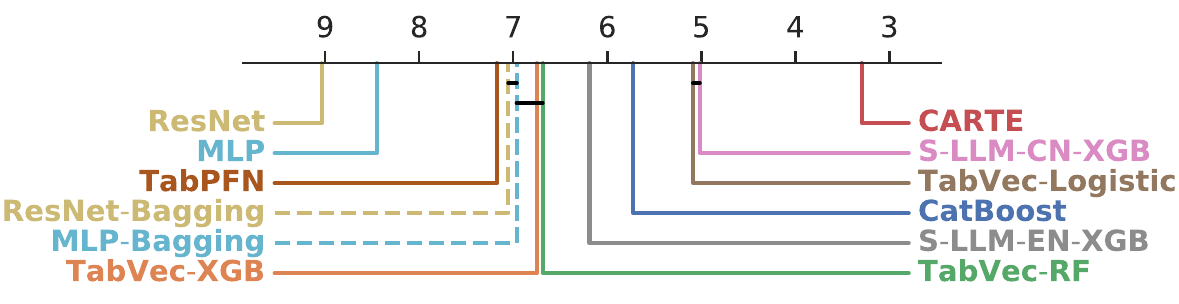}%
    \caption{\textbf{CARTE performs best for learning on single tables}. Learning curve on (a) regression and (b) classification tasks. Top: normalized score (1 is the best performer across all methods and train size for a dataset, and 0 the worst), averaged across datasets. Bottom: critical difference diagrams \citep{Terpilowski2019}, for all train sizes.
    \autoref{fig:comparison_all} gives critical difference diagram for all methods studied.\label{fig:learning_curve_and_critical_plot_regression}
    }
\end{figure*}

\subsection{Results on Single Tables}\label{subsec:results_singletables}
\paragraph{CARTE outperforms alternatives for learning on single tables} \autoref{fig:learning_curve_and_critical_plot_regression} compares the prediction performance of the multiple methods, summarizing the different datasets. We see that CARTE consistently outperforms alternatives across the different sample sizes, whether it is with normalized score\footnote{The calculation of the normalized score was adapted from \citet{grinsztajnWhyTreebasedModels2022}, in which the minimum score is fixed at a value $\rho$: $\rho=0$ for regression and $\rho=0.5$ for classification.} or critical difference diagrams based on the Conover post hoc test after the Friedman test to detect pairwise significance \citep{conover1999practical}. Another important point is that the bagging strategy used for CARTE also has positive impacts for neural network models: such a bagging with different train/validation splits for early stopping may be beneficial for deep learning in general. Appendix \ref{app:comparison_overall} gives comprehensive results of CARTE and 42 baselines.

\paragraph{CARTE is robust to missing values.}
When handling missing values, CARTE discards columns with the missing value in the graph construction step. For example, a data point with one missing value on a table with 10 columns would have nine leaf-nodes after the graph construction step. \autoref{tab:missing_values} compares the percentage drop in performance and its normalized scores (in comparison to \autoref{fig:learning_curve_and_critical_plot_regression}) of CARTE and several decision tree baselines that inherently handle the missing values. In the experiment, we randomly drop a proportion of features for each sample (train/test inclusive). The fraction of dropped columns are set as 0.1 (10 \% features dropped) and 0.3. The table shows that CARTE continues to outperform the baselines with smaller decrease in performance created by missing values.

\begin{table}[t]
\setlength{\tabcolsep}{5.909pt}
\small%
\begin{center}
\caption{\textbf{CARTE is robust to missing values.} Percentage drop in performance and its normalized scores with missing values in which a proportion (0.1 or 0.3) of features are randomly dropped.%
\label{tab:missing_values}
}
\begin{tabular}{p{.2078\columnwidth}llll}
\multicolumn{5}{l}{\hspace{-0.7em}\textbf{Percentage decrease created by missing values}\vspace{1mm}}\\%
\toprule
\multicolumn{1}{c}{\multirow{2}{*}{\textbf{Methods}}} & \multicolumn{4}{c}{\textbf{Train size (Missing fraction)}} \\
\cmidrule(r){2-5}
& \multicolumn{1}{c}{64 (0.1)} & \multicolumn{1}{c}{64 (0.3)} & \multicolumn{1}{c}{512 (0.1)} & \multicolumn{1}{c}{512 (0.3)} \\
\midrule
\rowcolor{gray!25}
CARTE      & 13.28\%          & \textbf{38.35\%}   & \textbf{10.19\%} & \textbf{24.42\%} \\
CatBoost   & 21.70\%          & 53.32\%            & 12.23\%          & 29.70\% \\
\rowcolor{gray!25}
TabVec-XGB & 15.11\%          & 51.27\%            & 12.61\%          & 30.35\% \\
TabVec-RF  & ~\textbf{~7.68\%}  & 44.43\%            & 12.77\%          & 29.79\% \\
\bottomrule
\\
\multicolumn{5}{l}{\hspace{-0.7em}\textbf{Normalized absolute score} (as in  \autoref{fig:learning_curve_and_critical_plot_regression})\vspace{1mm}}\\
\toprule
\rowcolor{gray!25}
CARTE      & \textbf{0.44}$_{\textbf{(0.20)}}$ & \textbf{0.29}$_{\textbf{(0.18)}}$ & \textbf{0.75}$_{\textbf{(0.12)}}$ & \textbf{0.61}$_{\textbf{(0.14)}}$ \\
CatBoost   & 0.31$_{\text{(0.22)}}$ & 0.17$_{\text{(0.17)}}$   & 0.65$_{\text{(0.15)}}$ & 0.50$_{\text{(0.15)}}$ \\
\rowcolor{gray!25}
TabVec-XGB & 0.19$_{\text{(0.20)}}$ & 0.11$_{\text{(0.15)}}$   & 0.65$_{\text{(0.17)}}$ & 0.50$_{\text{(0.17)}}$ \\
TabVec-RF  & 0.23$_{\text{(0.21)}}$ & 0.14$_{\text{(0.16)}}$   & 0.63$_{\text{(0.15)}}$ & 0.49$_{\text{(0.15)}}$ \\
\bottomrule
\end{tabular}
\end{center}
\end{table}

\paragraph{Computation time trade-off for CARTE} 

\autoref{tab:computation_time} shows the average computation time (in seconds) of the top-four baselines. The strong prediction performance of CARTE comes at a cost in computation time, and the gap increases with train size ($n$). This cost calls for further optimizations.

\begin{table}[t]
\begin{center}
\setlength{\tabcolsep}{3.1pt}
\small
\caption{\textbf{Computation time (in seconds) for top-four baselines} for preprocessing, training and testing, across the 51 datasets.%
\label{tab:computation_time}
}
\begin{tabular}{llll}
\toprule
\multicolumn{1}{c}{\multirow{2}{*}{\textbf{Methods}}} & \multicolumn{1}{c}{\multirow{2}{*}{\textbf{Preprocessing}}} & \multicolumn{1}{c}{\textbf{Learning}} & \multicolumn{1}{c}{\textbf{Learning}} \\
 & & \multicolumn{1}{c}{$n=64$} & \multicolumn{1}{c}{$n=512$} \\
\midrule
\rowcolor{gray!25}
CARTE         & ~~50.20$\pm$63.68   & 85.43$\pm$60.30 & 315.49$\pm$119.84 \\
CatBoost      & ~~~~~-                 & ~~0.98$\pm$1.19   & ~~~~1.05$\pm$1.06 \\
\rowcolor{gray!25}
TabVec-XGB    & ~~64.72$\pm$139.23  & ~~0.40$\pm$0.21   & ~~~~1.19$\pm$0.94 \\
S-LLM-XGB     & 207.87$\pm$361.56 & ~~0.87$\pm$0.71   & ~~~~3.49$\pm$1.79 \\
\bottomrule
\end{tabular}
\end{center}
\end{table}

\paragraph{Entity matching not required for CARTE}

The hypothesis to explain the good performance of CARTE is that has integrated information on many entities because it has been pretrained on the YAGO knowledge base. However, in a given downstream table, entities might appear written in a different way: for instance ``Londres'' instead of ``London''. This begs the question of whether CARTE transfers well useful information if the string representation of the entities differs. String matching is necessary for instance when using as features vector embeddings of the entries such as KEN embeddings \citep{cvetkov2023relational}.

\begin{figure}[b]
    \includegraphics[width=\linewidth]{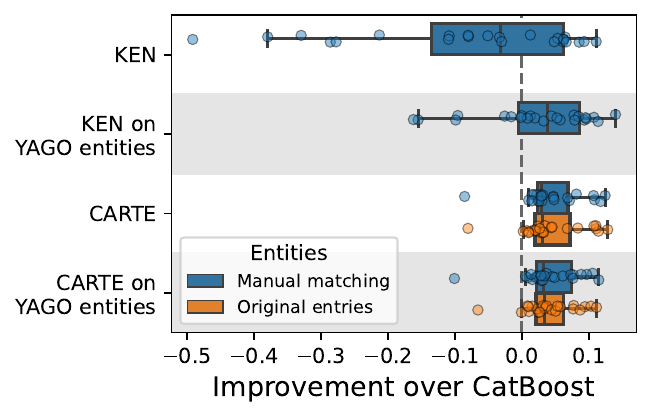}%
    \caption{\textbf{Entity matching not required for CARTE}, and downstream entities do not need to be in YAGO. We evaluate CARTE and KEN either on the full datasets, or on a reduced version of the datasets corresponding to entities present in YAGO. In addition, when entities are present in YAGO, we either match them to their canonical names in YAGO (blue) or keep the original names (orange). When KEN is used to enrich the dataset, CatBoost is used as the estimator, and entities without matching are replaced with missing values. Each point on the figure correspond to an improvement in performance with respect to Catboost without any enrichment. That KEN brings performance gains to CatBoost on YAGO entities confirms the added value of background information. Appendix \ref{app:matching} gives detailed results.
    \label{fig:entity_matching}
    }
\end{figure}

On four of our datasets (company employees, movies, US accidents, and US election), we performed manual entity matching of the entries to their corresponding YAGO entity. \autoref{fig:entity_matching} shows that while using KEN requires entity matching to have good performance, the string-level modeling in CARTE make its performance robust to entity variants: using manually matched entities or original entries. Ablations confirm the importance of string-level representations that also capture semantic similarity (Appendix \autoref{fig:ablation_attention}).

\begin{figure}[t]
    \centerline{\includegraphics[width=.75\linewidth]{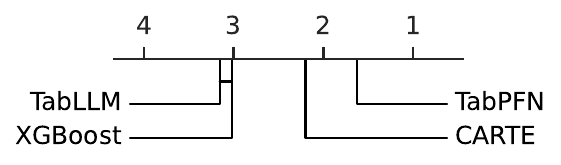}}%
    \caption{\textbf{Comparison to three baselines in TabLLM} \citep{hegselmannTabLLMFewshotClassification2023}. The datasets contain mostly numerical features or low-cardinality categorical columns. In such settings, TabPFN performs best, followed by CARTE, XGBoost, and TabLLM.
    \label{fig:cd_diagram_tabllm}}
\end{figure}

\paragraph{Comparison to TabLLM baselines} 
We compare CARTE to three baselines over nine datasets presented TabLLM \cite{hegselmannTabLLMFewshotClassification2023}. Compared to the datasets in \autoref{fig:learning_curve_and_critical_plot_regression}, the datasets in TabLLM contain higher fraction of numerical features with less cardinality of categorical columns (see \autoref{tab:data_specification}). \autoref{fig:cd_diagram_tabllm} gives the critical difference diagram of the baselines. TabPFN shows strength in such settings. Yet, CARTE can attain competitive performances although it is geared towards handling both numerical values and strings. Detailed results are analyzed in Appendix \ref{app:tabllm_all}.

\begin{table}[t]
\small%
\begin{center}
\caption{\textbf{Difference between this study's benchmark and TabLLM datasets.} Our benchmark datasets contain more categorical columns, in particular 
with higher cardinality ($|C|$).}
\label{tab:data_specification}
\vspace{0.1in}%
\begin{tabular}{lcc}
\toprule
\multicolumn{1}{c}{\multirow{2}{*}{\textbf{Characteristics}}} & \textbf{This study's} & \multirow{2}{*}{\textbf{TabLLM}} \\
& \textbf{benchmark} & \\
\midrule
Fraction of numerical cols.            & 0.194 & 0.613 \\
\rowcolor{gray!25}
Fraction of cols. with $|C|>10$   & 0.625 & 0.043 \\
Cardinality over data size          & 0.263 & 0.001 \\
\bottomrule
\end{tabular}
\end{center}
\end{table}

\subsection{Learning Across Multiple Tables}

\begin{figure}[t]
    \includegraphics[width=\linewidth]{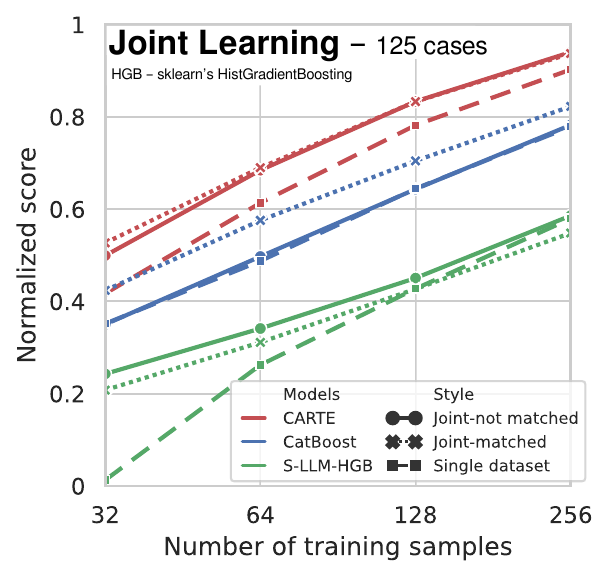}%
        \\%
    \includegraphics[width=\linewidth]{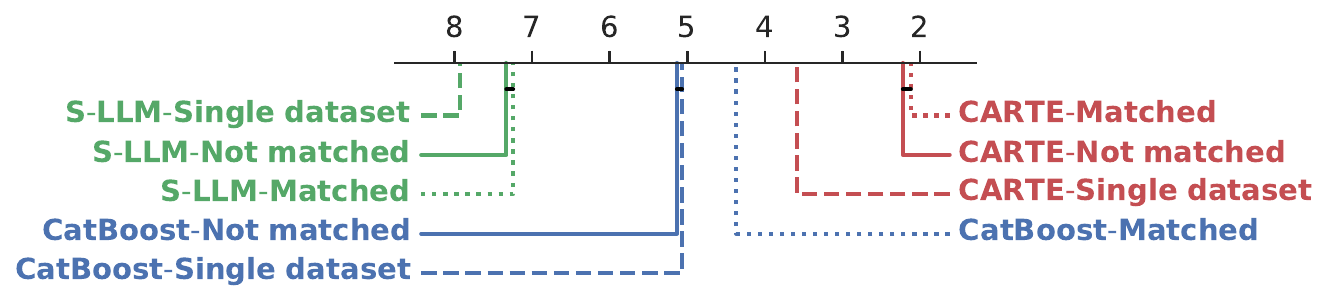}%
    \caption{\textbf{Schema matching not required for CARTE}, with consistent improvements through joint learning. We compare three scenarios (style) -- single (dashed lines): only the target tables; joint (full lines): the automated transfer learning without any manual operation; matched (dotted lines): transfer learning after manually matching the columns.
    \label{fig:learning_curve_multitable}
    }
\end{figure}

We investigate learning across multiple tables without explicit correspondences across columns. We use tables ``in the wild'': in our 51 datasets, we find groups covering the same general topic (bike prices, restaurant ratings) though they come from different sources (Appendix \ref{app:multitable}). In this setting, we can readily use CARTE and S-LLM approaches as they use an open-vocabulary representation of the column to embed entries (but only the EN, Embed Numerical, version of S-LLM, as the CN, Concat Numerical, needs correspondence for the numerical columns). As CatBoost natively deals with missing values, we use it by concatenating the datasets and adding missing values for mismatch columns. We also investigate manual matching of columns.

\paragraph{Schema matching not required for CARTE}
\autoref{fig:learning_curve_multitable} shows results for transfer learning across only two tables. We see that for all approaches transfer learning can help (the dashed line, representing the learning only on the target table, is below), but only CARTE provides consistent improvements without requiring manual column matching. For CatBoost, the matching (dotted line) is crucial, which is not the case for the other approaches. For S-LLM, the benefit of transfer drops rapidly with respect to the number of training samples. The results show that CARTE does not need schema matching, and it provides consistent improvements in the target table with transfer. Extended results of schema-matching can be found in Appendix \ref{app:schema-matching}.

\begin{figure}
    \includegraphics[width=\linewidth]{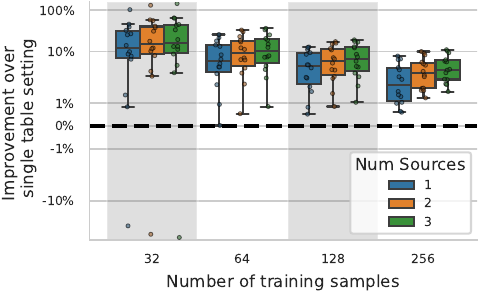}%
    \caption{\textbf{CARTE further benefits from additional source tables.}  }
    \label{fig:n_sources}
\end{figure}

\paragraph{Joint learning from multiple tables}
The difficulty for transfer learning may be finding good source tables. In \autoref{fig:n_sources}, we investigate bringing in more source tables, up to a total of 4 tables (1 target, and 3 sources, with a total of 245 cases). CARTE benefits from adding source tables: not only does the median performance improve but also, the lower bound in variability improves. In other words, more source tables give higher chances of finding a beneficial one, and thus the worst-case scenario becomes better.

\section{Discussion and Conclusion}

\paragraph{Strings and numbers in tables}

Our study touches on the importance of strings in tabular data. They are often overlooked in tabular machine learning (\autoref{tab:data_specification} shows how datasets are mostly numerical or low-cardinality categories), but central to database research, which focuses on discrete entries. Compared to most tabular-learning models (whether tree-based, or neural networks as TabPFN), the edge of CARTE is on the strings. String preprocessing in skrub's TableVectorizer also boosts baselines (\autoref{fig:comparison_all}).
Conversely, language models, as LLMs, focus on strings, enabling pretraining on huge corpora. They give great preprocessing of strings but must be combined with tree-based methods to handle numbers (S-LLM-CN-XGB in our benchmark). CARTE is is tailored to both strings and numbers.

\paragraph{An architecture that boosts performance} 

By using an architecture that models table entries not as a $i^\text{th}$ feature in a data matrix but as a function of its context (column names and neighboring entries) as well open-vocabulary embedding of strings, CARTE enables consistent representations of very different tables. This opens the door to pre-training across background tables, and fine-tuning to downstream tasks without matched entities or schema. Results show that after pretraining on a large knowledge base, the resulting model brings marked benefits to downstream analytic tasks, consistently outperforming a broad range of baselines both for learning on a single table or transfer learning on tables with imperfect correspondences.
It enables transfer from tables in the wild, a setting so far never studied.

\paragraph{Toward tabular foundation models}
Pre-training has been key to the wide application of deep learning on images and text. We hope that the ideas behind CARTE will bring these benefits to tabular learning, leading to tabular foundation models.
This will call for further improved architectures: optimizing for larger train-sizes; refining numerical representations with ideas of \citet{gorishniy2022embeddings}; leveraging more training information and expressive attention as in large language models; merging with the complementary meta-learning ideas of TabPFN \citep{hollmannTabPFNTransformerThat2023}, which shines on heavily-numerical tables.

\section*{Impact Statement}

This paper presents work whose goal is to advance the field of Machine Learning. There are many potential societal consequences of our work, none which we feel must be specifically highlighted here. The societal impact of a method depends on how it is used. We do note, however, that tabular data are central to fields such as healthcare which uses a lot codes and more or less normalized entities (\emph{e.g.,} ICD10 codes for diseases, medical informatics as a field has invested hugely on data integration). We thus hope that pretraining a tabular model on health data could provide value to this field, and in turn positive societal impact.

On another topic, we note that our model, CARTE, comes with additional computational cost compared to baselines. We do expect that further research and engineering will bring these costs down. However, our work opens the door to pre-trained models for tabular data, one day maybe foundation models. These models have led to a race for ever-increasing size, which comes with dire consequences in terms of energy and financial cost, carrying over to ecology and concentration of power.

\section*{Acknowledgments}
The authors acknowledge the support in part by the French Agence Nationale de la Recherche under Grant ANR-20-CHIA-0026 (LearnI).

We also would like to thank the ICML reviewers, who challenged us in a good way, leading to improved empirical study of CARTE, and hence a better manuscript.

\bibliography{example_paper,leo_bib}
\bibliographystyle{icml2024}

\appendix
\onecolumn

\section{Detailed Information on Training}

\subsection{Pretrained Model of CARTE}\label{app:pretrain}

\paragraph{Model specification and training details}
The model specification and training details were largely referenced from the work of \cite{devlinBERTPretrainingDeep2019}. We set $12$ attention layers, each consisting of $12$ multi-head attentions, and the hidden dimension was fixed to the same size as the inputs ($300$). The resulting model contains over $9.3$ million parameters. To run the pretraining, we selected $128$ entities with one additional positive, resulting in the batch size of $256$. The total number of steps for training was $1,000,000$, which approximately covers $40$ epochs with respect to YAGO entities. We use the AdamW optimizer accompanied by the cosine scheduler with $lr_{min}=5\times 10^{-6}$, $lr_{max}=1\times 10^{-4}$ and a warmup over the first $10,000$ steps. The dropout rate was fixed to $0.1$ and the gelu activation function was used.

\subsection{Details on Experiment Settings for Downstream Tasks}\label{app:exp_settings}

\paragraph{Single tables} To evaluate the performances of baselines on single tables, we focused on the setting with limited train-size for each table varying from $32$, $64$, $128$, $256$, $512$, $1,024$, and $2,048$; the rest of the remaining data were set as the test set. To find the optimal hyperparameters of the baselines, $5$-fold cross-validation over $100$ random search iteration were carried out on all the comparing methods except for CARTE and TabPFN. For CARTE, the same $5$-fold cross-validation, but only the grid-search over the learning rate was conducted. For TabPFN, we ran with the default values, as suggested in the paper. For detailed information the hyperparameter spaces of each method, please refer to the Hyperparmeter tuning paragraph below. The performance was recorded on $10$ different train/test splits, with the performance measure set as the $R^2$ score for regression and the Area Under Receiver Operating Curve (AUROC) for classification tasks.

\paragraph{Joint learning across multiple tables} The experiment settings for joint learning is almost the same as in the single table setting, except for minor details. The number of train-set on the target was varied across $32$, $64$, $128$, $256$, while the same split was set as in the case of single-tables to make the results comparable. In terms of hyperparameter optimization, CARTE only takes the best values obtained from the the single-table case (\autoref{sec:carte}). For other baselines, the same scheme for hyperparameter search was conducted.

\paragraph{Hyperparameter space} The hyperparameter tuning was done using grid search for CARTE, as we only tune the learning rate, and with random search for the baselines as these come with more than two hyperparameters to tune. The hyperparmeter spaces for XGBoost, HistGradientBoosting, RandomForest, Resnet, and MLP baselines are based on that used in \citet{grinsztajnWhyTreebasedModels2022}; for CatBoost we follow that used in the CatBoost paper \citep{dorogushCatboost2018}. For the baselines in joint learning across multiple tables, we employ an additional hyperparameter `fraction source', which denote the fraction of source data used for training. \autoref{tab:hyperparmeter_space} below summarizes the hyperparameter spaces for each of the estimators.

\begin{table}[!h]
\small
\begin{center}
\caption{Hyperparameter space for CARTE and baseline estimators.}
\label{tab:hyperparmeter_space}
\vspace{0.1in}
\begin{tabular}{lll}
\toprule
\multicolumn{1}{c}{\textbf{Methods}} & \multicolumn{1}{c}{\textbf{Parameters}} & \multicolumn{1}{c}{\textbf{Grid}} \\
\midrule
CARTE & Learning rate & [$2.5$, $5$, $7.5$] $\times$ [$1e^{-4}$, $1e^{-3}$] \\
\midrule
\multirow{6}{*}{CatBoost} & Max depth & UniformInt [$2$, $10$] \\
& Learning rate & LogUniform [$1e^{-5}$, $1$]\\
& Bagging temperature & Uniform [$0$, $1$]\\
& $l_{2}$-leaf regularization & LogUniform [$1$, $10$]\\
& One hot max size & UniformInt [$2$, $25$]\\
& Iterations & UniformInt [$400$, $1000$]\\
\hline
\multirow{10}{*}{XGBoost} & Num estimators & UniformInt [$50$, $1000$] \\
& Max depth & UniformInt [$2$, $10$]\\
& Learning rate & LogUniform [$1e^{-5}$, $1$]\\
& Min child weight & LogUniform [$1$, $100$]\\
& Subsample & Uniform [$0.5$, $1$]\\
& Colsample by level & Uniform [$0.5$, $1$]\\
& Colsample by tree & Uniform [$0.5$, $1$]\\
& Gamma & LogUniform [$1e^{-8}$, $7$]\\
& Lambda & LogUniform [$1$, $4$]\\
& Alpha & LogUniform [$1e^{-8}$, $100$]\\ 
\midrule
\multirow{5}{*}{HistGradientBoosting} & Learning rate & LogUniform [$1e^{-2}$, $10$] \\
& Max depth & [None, $2$, $3$, $4$]\\
& Max leaf nodes & NormalInt [$31$, $5$]\\
& Min samples leaf & NormalInt [$20$, $2$]\\
& $l_2$-regularization & LogUniform [$1e^{-6}$, $1e^{3}$]\\
\midrule
\multirow{6}{*}{RandomForest} & Num estimators & UniformInt [$50$, $250$] \\
& Max depth & [None, $2$, $3$, $4$]\\
& Max features & [sqrt, sqrt, log2, None, $0.1$, $0.2$, $0.3$, $0.4$, $0.5$, $0.6$, $0.7$, $0.8$, $0.9$]\\
& Min samples leaf & LogUniform [$1.5$, $50.5$]\\
& Bootstrap & [True, False]\\
& Min impurity decrease & [$0$, $0.01$, $0.02$, $0.05$]\\
\midrule
\multirow{9}{*}{ResNet} & Num layers & UniformInt [$1$, $8$]\\
& Layer size & UniformInt [$32$, $512$]\\
& Hidden factor & UniformInt [$1$, $3$]\\
& Hidden dropout & Uniform [$0$, $0.5$]\\
& Residual dropout & Uniform [$0$, $0.5$]\\
& Learning rate & LogUniform [$1e^{-5}$, $1e^{-2}$]\\
& Weight decay & LogUniform [$1e^{-8}$, $1e^{-2}$]\\
& Normalization & [batchnorm, layernorm]\\
& Batch size & [$16$, $32$]\\
\midrule
\multirow{6}{*}{MLP} & Num layers & UniformInt [$1$, $4$]\\
& Layer size & UniformInt [$16$, $1024$]\\
& Dropout & Uniform [$0$, $0.5$]\\
& Learning rate & LogUniform [$1e^{-5}$, $1e^{-2}$]\\
& Weight decay & LogUniform [$1e^{-8}$, $1e^{-2}$]\\
& Batch size & [$16$, $32$]\\
\midrule
\multirow{2}{*}{Ridge Regression} & Solver & [svd, cholesky, lsqr, sag]\\
& Alpha & LogUniform [$1e^{-5}$, $100$]\\
\midrule
\multirow{3}{*}{Logistic Regression} & Solver & [newton-cg, lbfgs, liblinear]\\
& Penalty & [none, $l_1$, $l_2$, elasticnet]\\
& C & LogUniform [$1e^{-5}$, $100$]\\
\midrule
\multirow{1}{*}{Baselines in joint learning} & Source fraction & Uniform [$0$, $1$]\\
\bottomrule
\vspace{0.4cm}
\end{tabular}
\end{center}
\end{table}

\subsection{Hardware Specifications}\label{app:hardware_specification}
The pretrained model for CARTE was trained on GPUs. For rest of our experiments, they were run on 32 cores of CPU and the hardware was chosen based on availability. 

\begin{description}[itemsep=1pt, leftmargin = 1.15cm]
    \item [\textbf{GPUs:}] NVIDIA V100 (32GB VRAM)
    \item [\textbf{CPUs:}] AMD EPYC 7742 64-Core Processor, AMD EPYC 7702 64-Core Processor (512GB RAM), Intel(R) Xeon(R) CPU E5-2660 v2, Intel(R) Xeon(R) Gold 6226R CPU (256GB RAM)
\end{description}

\subsection{Implementation of CARTE}
The implementation of CARTE will be available at \url{https://github.com/soda-inria/carte}.

\section{Data Description}\label{app:datasets}

\subsection{Data Preprocessing}
Only minimal data preprocessing were carried out in data preparation. For all datasets, we excluded columns that contained only one unique value or had missing values over half the size of the dataset.

\subsection{Datasets}
We provide detailed description of the datasets used in our experiment study. 
\begin{enumerate}
    \item \textbf{Anime Planet}\footnote{\url{https://www.kaggle.com/datasets/hernan4444/animeplanet-recommendation-database-2020}} This dataset contains information about anime scrapped from the website Anime-Planet. The task is to predict the average rating of the anime on this site.
    \item \textbf{Babies R Us} \citep{magellandata}\footnote{\url{http://pages.cs.wisc.edu/\textasciitilde anhai/data/784\_data/bikes/csv\_files/babies_r_us.csv}} Information of baby products scraped from the Babies R Us website. The task is to predict the price of baby products.    
    \item \textbf{Buy Buy Baby} \citep{magellandata}\footnote{\url{http://pages.cs.wisc.edu/\textasciitilde anhai/data/784\_data/bikes/csv\_files/buy_buy_baby.csv}} Information of baby products scraped from the Buy Buy Baby website. The task is to predict the price of baby products.    
    \item \textbf{Beer Ratings}\footnote{\url{https://www.kaggle.com/datasets/ruthgn/beer-profile-and-ratings-data-set}} The dataset contains tasting profiles and consumer reviews for 3197 unique beers from 934 different breweries. The task is to predict overall review ratings of different beers.
    \item \textbf{Bikedekho} \citep{magellandata}\footnote{\url{http://pages.cs.wisc.edu/\textasciitilde anhai/data/784\_data/bikes/csv\_files/bikedekho.csv}} Information on bikes and scooters from bikedekho website in India. The task is to predict the price of bikes.
    \item \textbf{Bikewale} \citep{magellandata}\footnote{\url{http://pages.cs.wisc.edu/\textasciitilde anhai/data/784\_data/bikes/csv\_files/bikewale.csv}} Information on bikes and scooters from bikewale website in India. The task is to predict the price of bikes.
    \item \textbf{Cardekho}\footnote{\url{https://www.kaggle.com/datasets/sukritchatterjee/used-cars-dataset-cardekho}} This dataset contains information on used cars, with their listing price in the websit Cardekho. The task is to predict the price.
    \item \textbf{Chocolate Bar Ratings}\footnote{\url{https://www.kaggle.com/datasets/rtatman/chocolate-bar-ratings}}Dataset containing information and expert rating on cocoa batches. The task is to predict the rating.
    \item \textbf{Clear Corpus} \citep{crossleyLarge2023}\footnote{\url{https://www.commonlit.org/blog/introducing-the-clear-corpus-an-open-dataset-to-advance-research-28ff8cfea84a}} Generic information about the reading passage excerpts for elementary school students. The task is to predict the readability of the excerpts. The text feature is the name of the book, not the excerpt.
    \item \textbf{Coffee Ratings}\footnote{\url{https://www.kaggle.com/datasets/hanifalirsyad/coffee-scrap-coffeereview}} Dataset scraped from coffeereview.com containing information on various coffees. The task is to predict the review ratings of the coffees.
    \item \textbf{Company Employees}\footnote{\url{https://www.kaggle.com/peopledatalabssf/free-7-million-company-dataset}} Information on companies with over $1,000$ employees. The task is to predict the number of employees of each company.
    \item \textbf{Employee remuneration and expenses earning over 75000}\footnote{\url{https://opendata.vancouver.ca/explore/dataset/employee-remuneration-and-expenses-earning-over-75000/information/?disjunctive.department&disjunctive.title}} Remuneration and expenses for employees earning over \$75,000 per year. The task is to predict the remuneration of employees.
    \item \textbf{Employee Salaries}\footnote{\url{https://openml.org/d/42125}} Information on salaries for employees of the Montgomery County, MD. The task is to predict the current annual salary range of the employees.
    \item \textbf{Fifa22 Players}\footnote{\url{https://www.kaggle.com/datasets/joebeachcapital/fifa-players}} Information on soccer players and their ability scores in Fifa22 game. The task is to predict the player's wage.
    \item \textbf{Filmtv Movies}\footnote{\url{https://www.kaggle.com/datasets/stefanoleone992/filmtv-movies-dataset/data}} Information of movies and ratings scraped from an Italian movie review website Filmtv Movies. The task is to predict the public vote on movies.
    \item  \textbf{Journal Score JCR} Scientific journals and their descriptive features from Journal Citation Reports. The task is to predict the impact factors of the journals.
    \item  \textbf{Journal Score SJR} Scientific journals and their descriptive features from Scimago journal rank. The task is to predict the H-index of journals.
    \item  \textbf{Japanese Anime}\footnote{\url{https://www.kaggle.com/datasets/dbdmobile/myanimelist-dataset}} List of Japanese animes and their relevant information. The task is to predict score for the animes.
    \item  \textbf{K-Drama}\footnote{\url{https://www.kaggle.com/datasets/noorrizki/top-korean-drama-list-1500}} List of korean drama and their basic information from mydramalist website. The task is to predict the score of the Korean dramas.
    \item  \textbf{Michelin}\footnote{\url{https://www.kaggle.com/datasets/ngshiheng/michelin-guide-restaurants-2021}} List of restaurants along with additional details curated from the Michelin Restaurants guide. The task is to predict the award of the restaurants.
    \item  \textbf{ML/DS Salaries}\footnote{\url{https://ai-jobs.net/salaries/download/salaries.csv}} salary and basic information of workers in machine learning and data science industry. The task is to predict the salary of workers.
    \item  \textbf{Movie Revenues}\footnote{\url{https://www.kaggle.com/rounakbanik/the-movies-dataset}} Metadata of movies released on or before July 2017. The task is to predict the range of the box-office revenues. 
    \item  \textbf{Museums}\footnote{\url{https://www.kaggle.com/datasets/markusschmitz/museums}} General information on the US museums. The task is to predict the revenues across the museums.
    \item  \textbf{Mydramalist}\footnote{\url{https://www.kaggle.com/datasets/rajchinagundi/mydramalist-complete-dataset}} General information on Asian drama scraped from mydramalist website. The task is to predict the ratings of Asian dramas.
    \item  \textbf{NBA Draft}\footnote{\url{https://www.kaggle.com/datasets/mattop/nba-draft-basketball-player-data-19892021}} Information on all NBA Draft picks from 1989-2021. The task is to predict the `value over replacement' of players.
    \item  \textbf{Prescription Drugs}\footnote{\url{https://data.ca.gov/uk/dataset/prescription-drugs-introduced-to-market}} The data contains new prescription drugs introduced to market in California with a Wholesale Acquisition Cost (WAC) that exceeding Medicare Part D. The task is to predict WAC at introduction.
    \item \textbf{Ramen ratings}\footnote{\url{https://www.kaggle.com/datasets/ankanhore545/top-ramen-ratings-2022}} The dataset contains ratings and characteristics of various ramens produced from multiple countries. The task is to predict the range of ratings of the ramens.
    \item  \textbf{Roger Ebert}\footnote{\url{https://github.com/gabrielcs/movie-ratings-prediction}} The dataset contains movies ratings by famous critic Rogert Ebert. The task is to predict the range of ratings.
    \item  \textbf{Rotten Tomatoes} \citep{magellandata}\footnote{\url{http://pages.cs.wisc.edu/~anhai/data/784_data/movies1/csv_files/rotten_tomatoes.csv}} Contain information on movies that can be found in Rotten Tomatoes movie rating website. The task is to predict the rating values of the movies.
    \item \textbf{Spotify}\footnote{\url{https://www.kaggle.com/datasets/maharshipandya/-spotify-tracks-dataset}} Generic information on Spotify tracks with some associated audio features. The task is to predict the popularity of the albums.
    \item \textbf{US Accidents}\footnote{\url{https://smoosavi.org/datasets/us_accidents}} Information of accidents in US cities between 2016 and 2020. From this dataset, two tasks are conducted: (1) the range of accident counts for the US cities (2) the severity of the reported accidents.
    \item \textbf{US Presidential} \citep{cvetkov2023relational} Voting statistics in the 2020 US presidential election along with information on US counties. The task is to predict the range of voting numbers across US counties.
    \item \textbf{Used Cars 24}\footnote{\url{https://www.kaggle.com/datasets/avikasliwal/used-cars-price-prediction}} Information on used cars. The task is to predict the price.
    \item \textbf{Used Cars Benz Italy}\footnote{\url{https://www.kaggle.com/datasets/bogdansorin/second-hand-mercedes-benz-registered-2000-2023-ita}} Dataset containing information on used cars sold in Italy. The task is to predict the price.
    \item  \textbf{UsedCars.com} Dataset containing information on used cars usedcars.com. The task is to predict the price.
    \item  \textbf{Used Cars Pakistan}\footnote{\url{https://www.kaggle.com/datasets/mustafaimam/used-car-prices-in-pakistan-2021}} Dataset containing information on used cars sold in Pakistan. The task is to predict the price.
    \item  \textbf{Used Cars Saudi Arabia}\footnote{\url{https://www.kaggle.com/datasets/turkibintalib/saudi-arabia-used-cars-dataset}} Dataset containing information on used cars sold in Saudi Arabia from the Syarah Website. The task is to predict the price of used cars.
    \item  \textbf{Videogame Sales}\footnote{\url{https://www.kaggle.com/datasets/gregorut/videogamesales}} This dataset contains a list of video games with sales greater than 100,000 copies (scrape of vgchartz.com). The task is to predict the global sales of the videogames.
    \item  \textbf{Whisky}\footnote{\url{https://whiskyanalysis.com/index.php/database/}} Basic and tasting information on whiskies form the whiskyanalysis.com. The task is to predict the range of meta critic of the whiskies.
    \item  \textbf{Wikiliq}\footnote{\url{https://www.kaggle.com/datasets/limtis/wikiliq-dataset}} Information on alcohol that can be found in the Wikiliq website. We conducted two tasks to predict the prices of (1) beer and (2) spirits. 
    \item  \textbf{Wina Poland}\footnote{\url{https://www.kaggle.com/datasets/skamlo/wine-price-on-polish-market}} Information about wines on the polish market. The task is to predict the price.
    \item  \textbf{Wine.com}\footnote{\url{https://www.kaggle.com/datasets/manyregression/updated-wine-enthusiast-review}} Information on wines scraped from the wine.com website. We conducted two tasks on prediction of (1) wine ratings and (2) wine prices.
    \item  \textbf{WineEnthusiasts}\footnote{\url{https://www.kaggle.com/datasets/manyregression/updated-wine-enthusiast-review}} Information about a wine and a taster from winemag.com. We conducted two tasks on prediction of (1) wine ratings and (2) wine prices.
    \item  \textbf{WineVivino}\footnote{\url{https://www.kaggle.com/datasets/joshuakalobbowles/vivino-wine-data}} Information about wine bottles scrapped from Vivino's website. We conducted two tasks on prediction of (1) wine ratings and (2) wine prices.
    \item  \textbf{Yelp}\footnote{\url{https://www.yelp.com/dataset}} The Yelp Open dataset for academic research. We extracted information on the restaurants from the original dataset. The task is to predict the range of stars for the restaurants. 
    https://www.yelp.com/dataset
    \item \textbf{Zomato}\footnote{\url{https://www.kaggle.com/datasets/anas123siddiqui/zomato-database?select=restaurant.csv}} Information and reviews of restaurants found in the zomato websites. The task is to predict the range of ratings for each restaurants.    
\end{enumerate}

\subsection{Datasets from multi-table experiments}\label{app:multitable}

For the multi-table experiments we extract from the list above groups of tables that are related to the same topics:
\begin{description}
    \item \textbf{Wine prices}: Wina Poland, WineEnthusiasts, WineVivino, Wine.com
    \item \textbf{Wine ratings}: WineEnthusiasts, WineVivino, Wine.com
    \item \textbf{Beers}: Beer Ratings, Wikiliq-Beer
    \item \textbf{Used Car}: Used Cars 24, Used Cars Benz Italy, UsedCars.com, Used Cars Pakistan, Used Cars Saudi Arabia
    \item \textbf{Films}: Filmtv Movies, Rotten Tomatoes
    \item \textbf{Dramas}: K-Drama, Mydramalist
    \item \textbf{Animes}: Anime Planet, Japanese Anime
    \item \textbf{Baby products}: Buy Buy Baby, Babies R Us
    \item \textbf{Bike sales}: Bikedekho, Bikewale
    \item \textbf{Employee remunerations}: Company Employees, Employee remuneration and expenses earning over 75000, ML/DS Salaries
    \item \textbf{Restaurant ratings}: Zomato, Michelin, Yelp
    \item \textbf{Journal scores}: Journal Score JCR, Journal Score SJR
\end{description}

\section{Extended Results}

\subsection{Performance Comparison of CARTE with 42 Baselines for Learning on Single Tables}\label{app:comparison_overall}

\begin{figure}[p!]
         \centering
         \textsf{(a) Regression}
         
         \includegraphics[width=\textwidth]{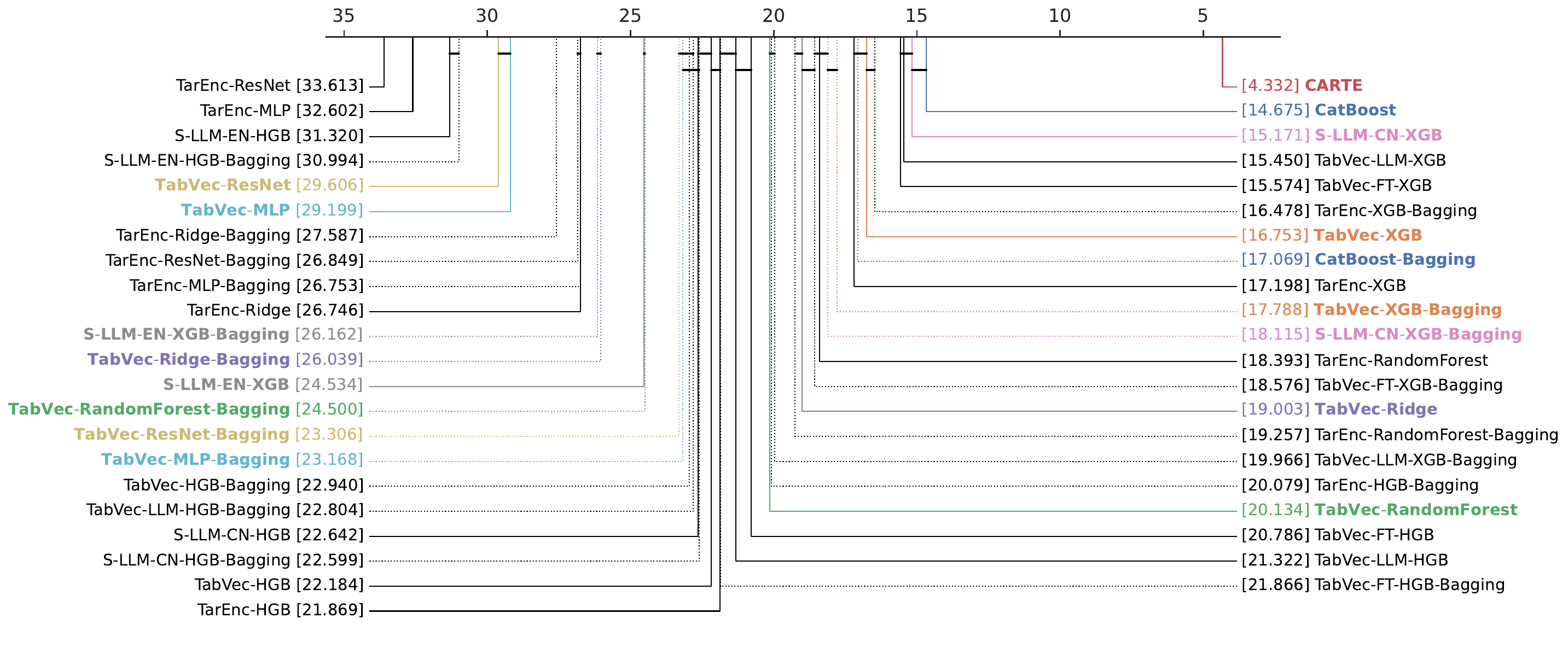}
         \bigskip%

         \centering
         \textsf{(b) Classification}
         
         \includegraphics[width=\textwidth]{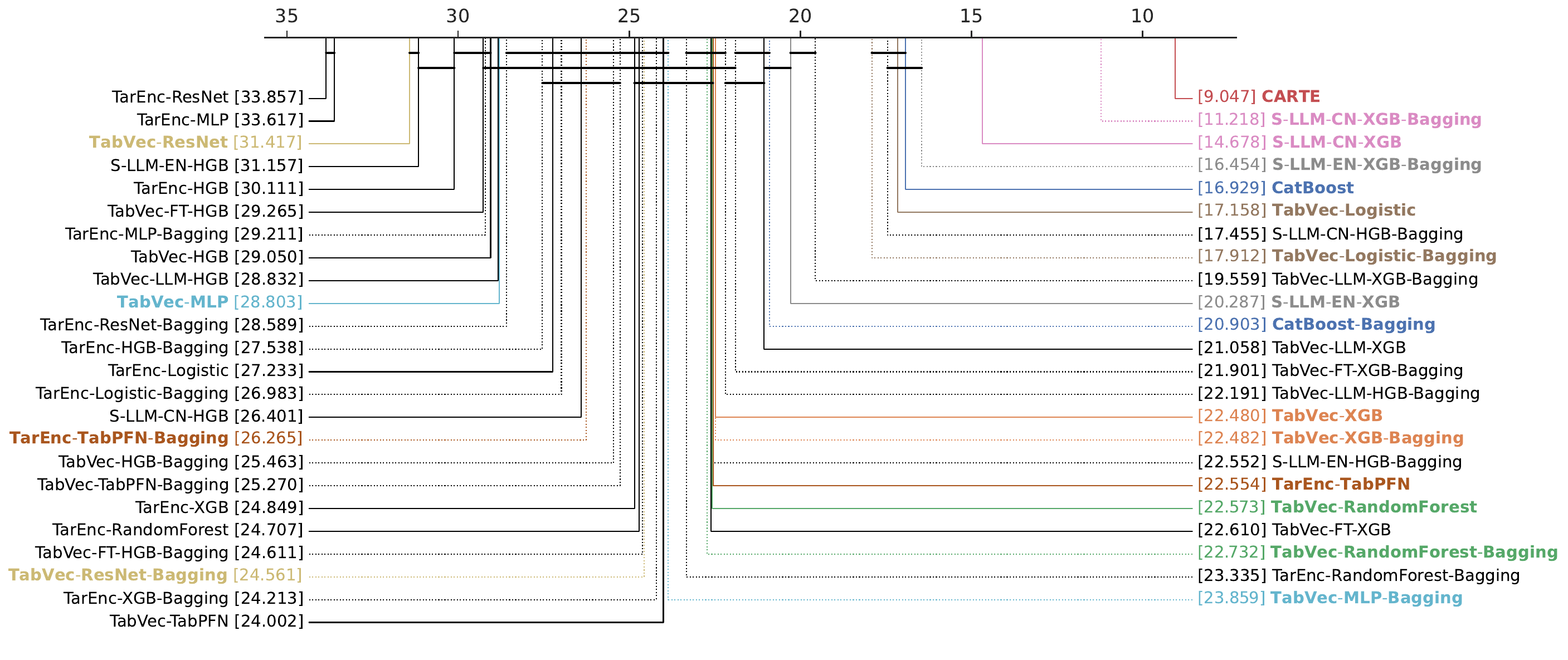}
        \caption{\textbf{Comparison of CARTE with 42 baselines on single tables.} The critical difference diagram for CARTE and 42 baselines for (a) regression (b) classification tasks. In addition to the methods in \autoref{fig:learning_curve_and_critical_plot_regression} (with the same coloring), we include additional baselines with HistGradientBoosting, target encoding, and external information from language models. The figure shows that CARTE attains the pronounced lead to all the baselines for both regression and classification tasks. Moreover, the bagging strategy brings larger benefits to neural networks compared to linear or ensembling baselines, which suggests an important setups for other deep learning architectures.
        \label{fig:comparison_all}}
\end{figure}

As an extension to the results shown in \autoref{subsec:results_singletables}, \autoref{fig:comparison_all} shows the overall comparison of CARTE with 42 baselines that additionally accounts for Scikit-Learn's HistGradientBoosting (\textbf{HGB}), target encoding for categorical variables \cite{micci2001preprocessing} (\textbf{TarEnc}), employing external information from language models of Fasttext (\textbf{FT}) and {\tt intfloat/e5-small-v2} \citep{wang2022text} (\textbf{LLM}), and the `\textbf{Bagging}' strategy. We see that CARTE attains the pronounced lead to all the baselines for both regression and classification tasks. Moreover, it is interesting to observe that the bagging strategy has positive impacts for neural network models, while the effect is limited on linear or ensembling baselines (tree-based models or TabPFN). This may hint that bagging with different train/validation splits is an important setups for other deep learning architectures, especially for limited train-sizes.

\begin{table}[!p]
\small%
\begin{center}
\caption{\textbf{Detailed results of TabLLM datasets.} Dataset specification and performance comparison among CARTE and three baselines presented in TabLLM \citep{hegselmannTabLLMFewshotClassification2023}. The datasets contain high fraction of numerical features or categorical columns with low cardinality. As observed from the comparison results, TabPFN performs the best followed by CARTE, XGBoost, and TabLLM.
\label{tab:tabllm}
}
\vspace{0.1in}
\begin{tabular}{ccc}
\multicolumn{3}{l}{\hspace{-0.7em}\textbf{TabLLM dataset specificiations}\vspace{1mm}}\\%
\toprule
\multicolumn{1}{c}{\multirow{2}{*}{\textbf{Datasets}}} & \multicolumn{1}{c}{\textbf{Fraction of}} & \multicolumn{1}{c}{\textbf{Average}} \\
& \multicolumn{1}{c}{\textbf{numerical columns}} & \multicolumn{1}{c}{\textbf{cardinality}} \\
\midrule
bank        & 0.40 & 38.44 \\
blood       & 1.00 & 0.00  \\
calhousing  & 1.00 & 0.00  \\
car         & 0.00 & 3.50  \\
creditg     & 0.33 & 3.79  \\
diabetes    & 1.00 & 0.00  \\
heart       & 0.45 & 2.67  \\
income      & 0.33 & 12.38 \\
jungle      & 1.00 & 0.00  \\
\bottomrule
\end{tabular}

\vspace{1mm}

\begin{tabular}{cclllll}
\\
\multicolumn{7}{l}{\hspace{-0.7em}\textbf{Performance comparison}\vspace{1mm}}\\%
\toprule
\multicolumn{1}{c}{\multirow{2}{*}{\textbf{Datasets}}} & \multicolumn{1}{c}{\multirow{2}{*}{\textbf{Methods}}} & \multicolumn{5}{c}{\textbf{Number of Shots}} \\ 
\cmidrule(r){3-7}
& & \multicolumn{1}{c}{\textbf{32}}  & \multicolumn{1}{c}{\textbf{64}}  & \multicolumn{1}{c}{\textbf{128}} & \multicolumn{1}{c}{\textbf{256}} & \multicolumn{1}{c}{\textbf{512}} \\
\midrule
\multirow{4}{*}{bank} & CARTE & \textbf{0.81$\pm$0.03} &  0.83$\pm$0.03 &  \textbf{0.87$\pm$0.04} &  0.89$\pm$0.03 &   0.90$\pm$0.01 \\
& XGBoost & 0.76$\pm$0.03 & 0.83$\pm$0.02 & 0.85$\pm$0.03 & 0.88$\pm$0.01 & 0.90$\pm$0.01 \\
& TabPFN & 0.76$\pm$0.03 & 0.82$\pm$0.03 & 0.86$\pm$0.02 & 0.89$\pm$0.00 & 0.90$\pm$0.00 \\
& TabLLM & 0.64$\pm$0.06 & 0.69$\pm$0.03 & 0.82$\pm$0.05 & 0.87$\pm$0.01 & 0.88$\pm$0.01 \\
\midrule
\multirow{4}{*}{blood} & CARTE & 0.68$\pm$0.01 &  0.68$\pm$0.01 &  0.72$\pm$0.02 &   0.72$\pm$0.0 &  0.71$\pm$0.01 \\
& XGBoost & 0.67$\pm$0.06 & 0.68$\pm$0.05 & 0.71$\pm$0.06 & 0.70$\pm$0.07 & 0.67$\pm$0.06 \\
& TabPFN & \textbf{0.70$\pm$0.04} & \textbf{0.73$\pm$0.04} & \textbf{0.75$\pm$0.04} & \textbf{0.76$\pm$0.04} & \textbf{0.76$\pm$0.03} \\
& TabLLM & 0.68$\pm$0.04 & 0.68$\pm$0.04 & 0.68$\pm$0.06 & 0.70$\pm$0.08 & 0.68$\pm$0.04 \\
\midrule
\multirow{4}{*}{calhousing} & CARTE & 0.79$\pm$0.02 &  0.83$\pm$0.03 &  0.85$\pm$0.04 &  0.87$\pm$0.05 &  0.89$\pm$0.05 \\
& XGBoost & 0.79$\pm$0.04 & 0.82$\pm$0.04 & 0.87$\pm$0.01 & 0.90$\pm$0.01 & 0.92$\pm$0.01 \\
& TabPFN & \textbf{0.85$\pm$0.03} & \textbf{0.89$\pm$0.01} & \textbf{0.91$\pm$0.01} & \textbf{0.92$\pm$0.00} & \textbf{0.93$\pm$0.00} \\
& TabLLM & 0.77$\pm$0.08 & 0.77$\pm$0.04 & 0.81$\pm$0.02 & 0.83$\pm$0.01 & 0.86$\pm$0.02 \\
\midrule
\multirow{4}{*}{car} & CARTE & 0.87$\pm$0.06 &  0.94$\pm$0.07 &  0.98$\pm$0.03 &  0.99$\pm$0.03 & 1.00$\pm$0.02 \\
& XGBoost & 0.82$\pm$0.03 & 0.91$\pm$0.02 & 0.95$\pm$0.01 & 0.98$\pm$0.01 & 0.99$\pm$0.01 \\
& TabPFN & \textbf{0.92$\pm$0.02} & \textbf{0.97$\pm$0.00} & \textbf{0.99$\pm$0.01} & \textbf{1.00$\pm$0.00} & 1.00$\pm$0.00 \\
& TabLLM & 0.91$\pm$0.02 & 0.96$\pm$0.02 & 0.98$\pm$0.01 & 0.99$\pm$0.00 & 1.00$\pm$0.00 \\
\midrule
\multirow{4}{*}{creditg} & CARTE & 0.67$\pm$0.03 &  0.68$\pm$0.01 &   0.70$\pm$0.02 &  0.75$\pm$0.01 &  \textbf{0.77$\pm$0.02} \\
& XGBoost & 0.66$\pm$0.03 & 0.67$\pm$0.06 & 0.68$\pm$0.02 & 0.73$\pm$0.02 & 0.75$\pm$0.03 \\
& TabPFN & 0.69$\pm$0.07 & 0.70$\pm$0.07 & \textbf{0.72$\pm$0.06} & 0.75$\pm$0.04 & 0.75$\pm$0.02 \\
& TabLLM & \textbf{0.72$\pm$0.06} & 0.70$\pm$0.07 & 0.71$\pm$0.07 & 0.72$\pm$0.03 & 0.72$\pm$0.02 \\
\midrule
\multirow{4}{*}{diabetes} & CARTE & 0.76$\pm$0.06 &  0.79$\pm$0.02 &  0.81$\pm$0.01 &  0.82$\pm$0.01 &   0.81$\pm$0.00 \\
& XGBoost & 0.69$\pm$0.08 & 0.73$\pm$0.05 & 0.78$\pm$0.05 & 0.80$\pm$0.03 & 0.80$\pm$0.01 \\
& TabPFN & \textbf{0.77$\pm$0.03} & \textbf{0.82$\pm$0.03} & \textbf{0.83$\pm$0.03} & \textbf{0.83$\pm$0.03} & 0.81$\pm$0.02 \\
& TabLLM & 0.68$\pm$0.04 & 0.73$\pm$0.03 & 0.79$\pm$0.04 & 0.78$\pm$0.02 & 0.78$\pm$0.04 \\
\midrule
\multirow{4}{*}{heart} & CARTE &  0.90$\pm$0.02 &  0.91$\pm$0.02 &  0.92$\pm$0.02 &  \textbf{0.93$\pm$0.01} &  \textbf{0.93$\pm$0.01} \\
& XGBoost & 0.88$\pm$0.04 & 0.91$\pm$0.01 & 0.91$\pm$0.01 & 0.90$\pm$0.01 & 0.92$\pm$0.01 \\
& TabPFN & \textbf{0.91$\pm$0.02} & \textbf{0.92$\pm$0.02} & 0.92$\pm$0.02 & 0.92$\pm$0.01 & 0.92$\pm$0.02 \\
& TabLLM & 0.87$\pm$0.06 & 0.91$\pm$0.01 & 0.90$\pm$0.01 & 0.92$\pm$0.01 & 0.92$\pm$0.01 \\
\midrule
\multirow{4}{*}{income} & CARTE & 0.84$\pm$0.09 &  0.84$\pm$0.02 &  0.85$\pm$0.03 &  0.87$\pm$0.01 &  0.88$\pm$0.01 \\
& XGBoost & 0.79$\pm$0.03 & 0.82$\pm$0.02 & 0.84$\pm$0.01 & 0.87$\pm$0.01 & 0.88$\pm$0.00 \\
& TabPFN & 0.80$\pm$0.04 & 0.82$\pm$0.04 & 0.84$\pm$0.01 & 0.86$\pm$0.01 & 0.87$\pm$0.01 \\
& TabLLM & 0.84$\pm$0.01 & 0.84$\pm$0.02 & \textbf{0.86$\pm$0.01} & 0.87$\pm$0.00 & \textbf{0.89$\pm$0.01} \\
\midrule
\multirow{4}{*}{jungle} & CARTE & 0.71$\pm$0.03 &   0.80$\pm$0.02 &  0.81$\pm$0.02 &  0.86$\pm$0.02 &   0.90$\pm$0.02 \\
& XGBoost & 0.78$\pm$0.03 & 0.81$\pm$0.02 & 0.84$\pm$0.02 & 0.87$\pm$0.01 & 0.91$\pm$0.01 \\
& TabPFN & 0.78$\pm$0.02 & 0.81$\pm$0.01 & 0.84$\pm$0.01 & \textbf{0.88$\pm$0.01} & 0.91$\pm$0.00 \\
& TabLLM & 0.71$\pm$0.02 & 0.78$\pm$0.02 & 0.81$\pm$0.02 & 0.84$\pm$0.01 & 0.89$\pm$0.01 \\
\bottomrule
\end{tabular}
\end{center}
\end{table}

\subsection{Detailed Results for TabLLM Datasets}\label{app:tabllm_all}
\autoref{tab:tabllm} shows the dataset specifications and detailed results on performance comparison between CARTE and baselines presented in \citet{hegselmannTabLLMFewshotClassification2023}. The datasets generally contain high fraction of numerical features (four datasets) or categorical columns with low cardinality (eight datasets). In such settings, TabPFN tends to outperform other methods. For the dataset `bank', however, CARTE outperforms other baselines. In particular, the dataset is in line with the 51 datasets that contain both numerical and categorical features with relatively high cardinality of the latter. In a sense, CARTE is in the middle of both TabPFN (suitable for numerical features) and TabLLM (representing information as tokens), with an attentional architecture that has been designed to handle both numerical values and strings.

\subsection{Ablation Study on the Components of CARTE}\label{app:ablation_attention}
\begin{figure}[!h]
\begin{center}
     \begin{subfigure}{0.497\textwidth}
         \centering
         \caption*{\textsf{(a) Regression}}
         \includegraphics[width=\textwidth]{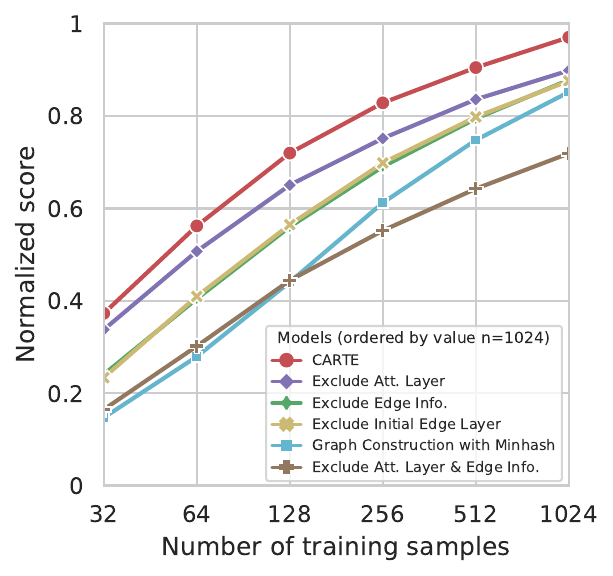}
     \end{subfigure}
     \begin{subfigure}{0.497\textwidth}
         \centering
         \caption*{\textsf{(b) Classification}}
         \includegraphics[width=\textwidth]{plots/ds_ablation_lc_reg.pdf}
     \end{subfigure}
\end{center}     
        \caption{\textbf{Ablation on various components of CARTE.} The learning curves from the train size of 32 up to 1,024 on various cases excluding or switching the components. Each are crucial for gaining the performance of CARTE. In particular, there is significant decrease with the exclusion of edge information and the attention layer, which are essential for leveraging context within a given table. Moreover, the result with Minhash \citep[A string-level representation without any semantic content,][]{cerdaEncodingHighcardinalityString2022} shows that language models are crucial for effectively using external information.}
        \label{fig:ablation_attention}
\end{figure}

To study the effect of various components of CARTE, we conducted additional experiments in which we exclude or change the associated components. \autoref{fig:ablation_attention} shows the learning curves from the train size of 32 up to 1,024 for each case of excluding or switching the components. For the graph construction with Minhash, we change the feature initialization step with Skrub's Minhash encoder \cite{skrub2024}, which encode string categorical features by applying the MinHash method to n-gram decompositions of strings. The figure shows that each are crucial for gaining the performance of CARTE. In particular, it is interesting to observe the significant decrease with the exclusion of edge information and the attention layer. Since both are essential for leveraging context within a given table, it implies that capturing context is pivotal for attaining the strong performances in predictions. Moreover, the performance gap between CARTE and Minhash confirm that string-level models, that also capture semantic similarity, is important and the use of language models are pivotal for effectively using external information, especially when information given in the table is limited.

\subsection{The Effect of Oversmoothing}\label{app:oversmoothing}

\begin{figure}[h]
    \begin{minipage}{.48\linewidth}
        \caption{\textbf{The effect of oversmoothing}: comparing prediction from representations extracted from the GNN from the 2\textsuperscript{nd}, 4\textsuperscript{th}, 8\textsuperscript{th}, and 12\textsuperscript{th} layers, with that build from the first layer. We see that the deeper we go in the GNN, the less useful the representation is for downstream task. We interpret this as an effect of oversmoothing.
        \label{fig:oversmoothing}
    }
    \end{minipage}%
    \hfill%
    \begin{minipage}{.48\linewidth}
        \includegraphics[width=\linewidth]{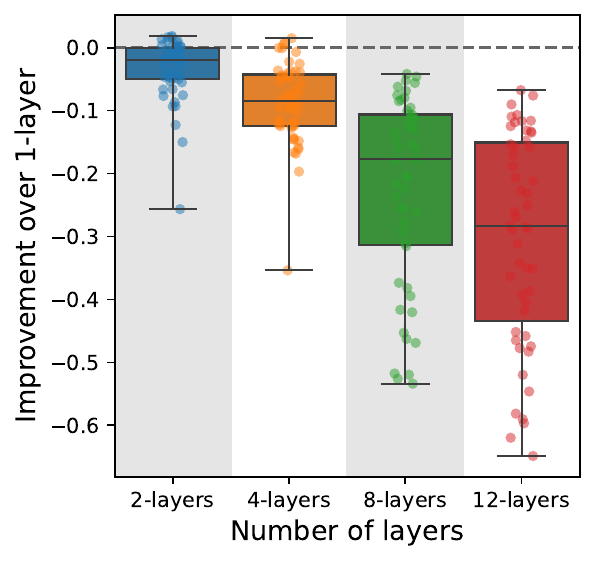}%
    \end{minipage}

\end{figure}

\autoref{fig:oversmoothing} show that representations extracted from deeper layers of the Graph Neural Networks (GNNs) are less useful for prediction on downstream tasks\footnote{Here, they are used inside a HistGradientBoosting predictor from scikit-learn.}. We interpret this as an effect of oversmoothing, a well known problem in GNNs \citep{chen2020measuring, rusch2023survey}.

\newpage

\subsection{Details of the Entity Matching Experiment}\label{app:matching}

The experiments were conducted with the same experiment settings as that of the singletable experiments. Table \ref{tab:matching} gives the specific results behind each dataset used in the entity matching experiment \autoref{fig:entity_matching}. The specific datasets are
\begin{itemize}\itemsep-0.2em
\item CE = Company employees : 32\% of the companies matched to YAGO
\item MV = movie revenues : 84\% of the movies matched to YAGO
\item US-Acc = US accidents : 67\% of the cities matched to YAGO
\item US-Elec = US elections : 98\% of the counties matched to YAGO
\end{itemize}

\begin{table}[!h]\footnotesize
\caption{\textbf{Detailed results of the entity matching experiment}: individual scores on each dataset. The abbreviations are as follows: O-Original entries, M-Matched entries, R-Reduced dataset, and F-Full dataset.%
\label{tab:matching}
}
\vspace{0.1in}
\begin{center}
    \begin{tabular}{lllll}
    \toprule
    {} &     \multicolumn{1}{c}{\textbf{CatBoost-MR}} &     \multicolumn{1}{c}{\textbf{CatBoost-MF}} &     \multicolumn{1}{c}{\textbf{CatBoost-OR}} &     \multicolumn{1}{c}{\textbf{CatBoost-OF}} \\
    \midrule
    CE-32        &   0.673$\pm$0.036 &   0.672$\pm$0.063 &   0.683$\pm$0.052 &   0.668$\pm$0.062 \\
    CE-64        &   0.707$\pm$0.021 &    0.72$\pm$0.013 &   0.702$\pm$0.025 &   0.718$\pm$0.017 \\
    CE-128       &   0.734$\pm$0.007 &   0.739$\pm$0.024 &   0.731$\pm$0.011 &    0.745$\pm$0.01 \\
    CE-256       &   0.739$\pm$0.008 &   0.747$\pm$0.014 &    0.74$\pm$0.009 &   0.749$\pm$0.008 \\
    CE-512       &   0.744$\pm$0.005 &   0.758$\pm$0.005 &   0.744$\pm$0.005 &   0.758$\pm$0.004 \\
    CE-1024      &   0.752$\pm$0.006 &   0.763$\pm$0.004 &   0.752$\pm$0.006 &   0.764$\pm$0.003 \\
    \midrule
    MV-32        &     0.4$\pm$0.058 &   0.398$\pm$0.049 &   0.394$\pm$0.058 &   0.403$\pm$0.042 \\
    MV-64        &   0.436$\pm$0.043 &   0.449$\pm$0.045 &   0.426$\pm$0.059 &   0.453$\pm$0.035 \\
    MV-128       &   0.484$\pm$0.027 &   0.492$\pm$0.028 &   0.482$\pm$0.018 &   0.495$\pm$0.019 \\
    MV-256       &   0.511$\pm$0.011 &   0.515$\pm$0.017 &    0.51$\pm$0.017 &   0.523$\pm$0.012 \\
    MV-512       &   0.545$\pm$0.007 &    0.55$\pm$0.007 &   0.545$\pm$0.009 &   0.552$\pm$0.008 \\
    MV-1024      &   0.559$\pm$0.007 &   0.574$\pm$0.007 &   0.563$\pm$0.005 &   0.573$\pm$0.005 \\
    \midrule
    US-Acc-32    &  -0.016$\pm$0.063 &  -0.018$\pm$0.064 &  -0.023$\pm$0.076 &   -0.02$\pm$0.058 \\
    US-Acc-64    &   0.007$\pm$0.055 &   -0.01$\pm$0.069 &   0.028$\pm$0.036 &  -0.023$\pm$0.087 \\
    US-Acc-128   &   0.082$\pm$0.026 &   0.055$\pm$0.029 &   0.084$\pm$0.028 &   0.057$\pm$0.026 \\
    US-Acc-256   &   0.129$\pm$0.018 &   0.089$\pm$0.031 &   0.129$\pm$0.015 &   0.089$\pm$0.025 \\
    US-Acc-512   &    0.163$\pm$0.02 &    0.12$\pm$0.022 &   0.163$\pm$0.021 &    0.121$\pm$0.02 \\
    US-Acc-1024  &   0.214$\pm$0.007 &    0.157$\pm$0.01 &   0.217$\pm$0.005 &   0.155$\pm$0.009 \\
    \midrule
    US-Elec-32   &    0.31$\pm$0.133 &   0.318$\pm$0.142 &    0.34$\pm$0.118 &   0.285$\pm$0.161 \\
    US-Elec-64   &   0.433$\pm$0.062 &   0.441$\pm$0.068 &   0.449$\pm$0.038 &   0.445$\pm$0.056 \\
    US-Elec-128  &   0.512$\pm$0.019 &    0.505$\pm$0.02 &    0.511$\pm$0.02 &     0.51$\pm$0.02 \\
    US-Elec-256  &   0.547$\pm$0.009 &   0.543$\pm$0.011 &   0.546$\pm$0.009 &   0.544$\pm$0.011 \\
    US-Elec-512  &   0.572$\pm$0.007 &    0.571$\pm$0.01 &    0.57$\pm$0.009 &   0.571$\pm$0.008 \\
    US-Elec-1024 &   0.586$\pm$0.004 &   0.586$\pm$0.006 &   0.586$\pm$0.005 &   0.587$\pm$0.005 \\
    \bottomrule
    \end{tabular}

    \vspace{5mm}

    \begin{tabular}{lllllll}
    \toprule
    {} &       \multicolumn{1}{c}{\textbf{CARTE-MR}} &       \multicolumn{1}{c}{\textbf{CARTE-MF}} &       \multicolumn{1}{c}{\textbf{CARTE-OR}} &       \multicolumn{1}{c}{\textbf{CARTE-OF}} &         \multicolumn{1}{c}{\textbf{KEN-R}} &         \multicolumn{1}{c}{\textbf{KEN-F}} \\
    \midrule
    CE-32        &  0.699$\pm$0.023 &   0.69$\pm$0.034 &  0.693$\pm$0.023 &  0.692$\pm$0.029 &   0.518$\pm$0.11 &  0.459$\pm$0.119 \\
    CE-64        &  0.729$\pm$0.019 &  0.744$\pm$0.025 &   0.733$\pm$0.01 &  0.747$\pm$0.022 &  0.612$\pm$0.077 &  0.434$\pm$0.284 \\
    CE-128       &  0.755$\pm$0.007 &  0.763$\pm$0.012 &   0.755$\pm$0.01 &  0.763$\pm$0.014 &  0.708$\pm$0.019 &  0.409$\pm$0.305 \\
    CE-256       &  0.762$\pm$0.009 &   0.776$\pm$0.01 &  0.763$\pm$0.008 &  0.781$\pm$0.007 &   0.738$\pm$0.02 &  0.368$\pm$0.812 \\
    CE-512       &  0.773$\pm$0.011 &  0.785$\pm$0.007 &  0.778$\pm$0.012 &  0.789$\pm$0.008 &  0.757$\pm$0.006 &  0.267$\pm$0.494 \\
    CE-1024      &  0.783$\pm$0.008 &  0.793$\pm$0.006 &   0.787$\pm$0.01 &  0.798$\pm$0.006 &  0.772$\pm$0.008 &  0.486$\pm$0.301 \\
    \midrule
    MV-32        &    0.3$\pm$0.057 &  0.313$\pm$0.083 &  0.329$\pm$0.066 &  0.322$\pm$0.095 &  0.301$\pm$0.057 &  0.318$\pm$0.033 \\
    MV-64        &  0.452$\pm$0.044 &  0.471$\pm$0.027 &  0.458$\pm$0.025 &  0.461$\pm$0.038 &   0.42$\pm$0.035 &  0.369$\pm$0.055 \\
    MV-128       &  0.521$\pm$0.022 &  0.523$\pm$0.022 &  0.519$\pm$0.023 &  0.515$\pm$0.018 &  0.493$\pm$0.024 &   0.384$\pm$0.08 \\
    MV-256       &  0.556$\pm$0.022 &   0.562$\pm$0.02 &  0.554$\pm$0.021 &  0.555$\pm$0.014 &  0.543$\pm$0.014 &  0.464$\pm$0.089 \\
    MV-512       &  0.594$\pm$0.013 &  0.597$\pm$0.013 &  0.595$\pm$0.014 &  0.595$\pm$0.011 &  0.589$\pm$0.012 &   0.517$\pm$0.04 \\
    MV-1024      &   0.62$\pm$0.008 &  0.622$\pm$0.008 &   0.62$\pm$0.008 &  0.618$\pm$0.009 &  0.616$\pm$0.007 &  0.544$\pm$0.032 \\
    \midrule
    US-Acc-32    &  0.061$\pm$0.054 &  0.053$\pm$0.055 &  0.054$\pm$0.067 &  0.048$\pm$0.045 &  0.062$\pm$0.094 &  0.047$\pm$0.045 \\
    US-Acc-64    &  0.112$\pm$0.057 &  0.114$\pm$0.046 &  0.122$\pm$0.056 &  0.105$\pm$0.051 &  0.146$\pm$0.038 &   0.051$\pm$0.09 \\
    US-Acc-128   &   0.155$\pm$0.06 &  0.136$\pm$0.053 &   0.16$\pm$0.058 &   0.14$\pm$0.051 &  0.175$\pm$0.025 &  0.117$\pm$0.026 \\
    US-Acc-256   &  0.225$\pm$0.024 &  0.197$\pm$0.023 &   0.232$\pm$0.02 &    0.2$\pm$0.024 &  0.225$\pm$0.029 &  0.152$\pm$0.014 \\
    US-Acc-512   &  0.278$\pm$0.008 &   0.237$\pm$0.01 &  0.275$\pm$0.015 &  0.235$\pm$0.014 &   0.27$\pm$0.012 &  0.173$\pm$0.029 \\
    US-Acc-1024  &   0.303$\pm$0.01 &  0.263$\pm$0.008 &  0.304$\pm$0.008 &  0.265$\pm$0.012 &  0.298$\pm$0.004 &  0.205$\pm$0.006 \\
    \midrule
    US-Elec-32   &  0.387$\pm$0.082 &  0.387$\pm$0.083 &   0.393$\pm$0.08 &  0.393$\pm$0.082 &  0.149$\pm$0.193 &  0.209$\pm$0.159 \\
    US-Elec-64   &  0.467$\pm$0.031 &  0.467$\pm$0.032 &  0.465$\pm$0.021 &  0.465$\pm$0.022 &  0.432$\pm$0.089 &  0.454$\pm$0.073 \\
    US-Elec-128  &   0.52$\pm$0.021 &   0.52$\pm$0.021 &   0.52$\pm$0.022 &   0.52$\pm$0.022 &  0.564$\pm$0.038 &  0.571$\pm$0.035 \\
    US-Elec-256  &  0.552$\pm$0.011 &  0.553$\pm$0.011 &  0.546$\pm$0.013 &  0.546$\pm$0.013 &  0.625$\pm$0.019 &  0.628$\pm$0.011 \\
    US-Elec-512  &  0.586$\pm$0.008 &  0.586$\pm$0.008 &   0.58$\pm$0.007 &  0.581$\pm$0.007 &  0.667$\pm$0.006 &   0.664$\pm$0.01 \\
    US-Elec-1024 &  0.615$\pm$0.007 &  0.615$\pm$0.007 &  0.612$\pm$0.007 &  0.613$\pm$0.007 &    0.7$\pm$0.009 &  0.697$\pm$0.005 \\
    \bottomrule
    \end{tabular}
\end{center}
\end{table}

\subsection{Schema-matching Results on Joint Learning Across Multiple Tables}\label{app:schema-matching}

\begin{figure}[h]
    \begin{minipage}{.48\linewidth}
        \caption{\textbf{Performance comparison of CARTE with and without schema-matching}: The figures portrays a direct comparison of performance between CARTE with and without schema-matching over 275 different cases in number of source data. A point below the diagonal line indicate better performance of the method in x-axis. We see that the dots align along the diagonal line, showing similar performance both approaches of CARTE on joing learning (\emph{p}-value of 0.728). The results bolster no schema-matching for CARTE.
        \label{fig:schema_matching}
    }
    \end{minipage}%
    \hfill%
    \begin{minipage}{.48\linewidth}
        \includegraphics[width=\linewidth]{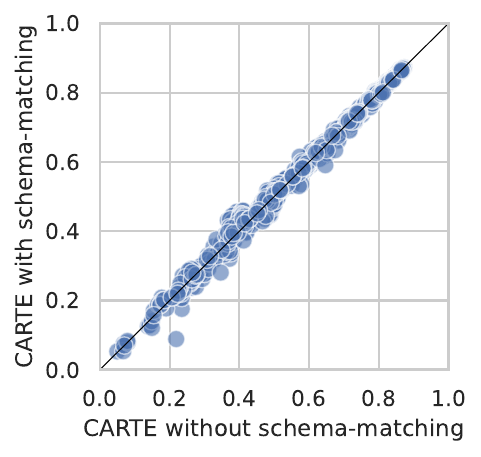}%
    \end{minipage}

\end{figure}

\autoref{fig:schema_matching} gives a direct comparison of performance between CARTE with and without schema-matching over 275 different cases in number of source data (ranging from one source to five sources). Each point represents a comparison of the average score over a dataset with different train/test split for a given size of train data. If a point is located below the diagonal line, it indicates a higher performance of x-axis. The figures shows that dots align along the diagonal line, indicating similar performance between CARTE with and without schema-matching (also with \emph{p}-value of 0.728 for two-sided t-test on difference in means). The results suggest schema-matching is not required for CARTE on transfer across multiple tables.

\end{document}